\documentclass{article}

 \usepackage[final]{corl_2020} %

\usepackage[utf8]{inputenc} %
\usepackage[T1]{fontenc}    %
\usepackage{hyperref}       %
\usepackage{url}            %
\usepackage{booktabs}       %
\usepackage{amsfonts}       %
\usepackage{nicefrac}       %
\usepackage{microtype}      %
\usepackage{amsmath}
\usepackage{textcomp}

\usepackage{microtype}
\usepackage{subfig}
\usepackage{graphicx}
\usepackage{booktabs} %
\usepackage{multirow}
\usepackage{amsfonts}
\usepackage{wrapfig}
\usepackage{subfig}
\usepackage{siunitx}
\usepackage{dcolumn}
\usepackage{wrapfig}
\usepackage[ruled]{algorithm2e}

\usepackage{graphicx}
\usepackage{float}

\newcommand{\ts}{\tilde{s}}
\newcommand{\ta}{\tilde{a}}
\newcommand{\tr}{\tilde{r}}
\newcommand{\tV}{\tilde{V}}

\makeatletter 
\newcommand\figcaption{\def\@captype{figure}\caption} 
\newcommand\tabcaption{\def\@captype{table}\caption} 
\makeatother

\title{Contrastive Variational Reinforcement Learning \\for Complex Observations}

\author{
  Xiao Ma, Siwei Chen, David Hsu, Wee Sun Lee\\
  National University of Singapore\\ 
  \texttt{\{xiao-ma,siwei-15,leews,dyhsu\}@comp.nus.edu.sg} \\
}

\begin{document}
\maketitle

\begin{abstract}
Deep reinforcement learning (DRL) has achieved significant success in various robot tasks: manipulation, navigation, \textit{etc.}\,. However, complex visual observations in natural environments remains a major challenge. This paper presents \textit{Contrastive Variational Reinforcement Learning} (CVRL), a model-based method that tackles complex visual observations in  DRL.  CVRL learns a contrastive variational world model discriminatively by maximizing the mutual information between latent states and observations, through \textit{contrastive learning}. It avoids modeling the complex observation space unnecessarily, as the commonly used generative observation model often does,  and is significantly more robust. We evaluated CVRL on challenging RL benchmark tasks that require continuous control. CVRL achieved comparable performance with  state-of-the-art model-based DRL methods  on  standard Mujoco tasks. It significantly outperformed them on \textit{Natural} Mujoco tasks and a robot box-pushing task with complex observations, \textit{e.g.}, dynamic shadows. The CVRL code is available publicly at \url{https://github.com/Yusufma03/CVRL}.

\end{abstract}


\section{Introduction}

Model-free reinforcement learning (MFRL) has achieved great success in game playing~\citep{mnih2013playing,silver2017mastering}, robot navigation~\citep{levine2016end,kahn2018self} and etc. 
However, extending existing RL methods to real-world environments remains challenging, because they require long-horizon reasoning with the low-dimensional useful features, \textit{e.g.}, the position of a robot, embedded in high-dimensional complex observations, \textit{e.g.}, visually rich images. Consider a four-legged mini-cheetah robot~\citep{bosworth2015super} navigating on the campus. To determine the traversable path, the robot must extract the relevant geometric features that coexist with irrelevant variable backgrounds, such as the moving pedestrians, paintings on the wall, etc.

Model-based RL (MBRL), in contrast to the model-free methods, reasons a world model trained by \textit{generative} learning and greatly improves the sample efficiency of the model-free methods~\citep{allen1983planning,basye1992decision,ha2018recurrent}. Recent MBRL methods learn compact latent world models from high-dimensional visual inputs with \textit{Variational Autoencoders} (VAEs)~\citep{kingma2013auto} by optimizing the \textit{evidence lower bound} (ELBO) of an observation sequence~\citep{igl2018deep,hafner2018planet}. However, learning a generative model under complex observations is challenging. VAEs learn the correspondence between observation $o_t$ and latent state $s_t$ by maximizing the conditional observation likelihood $p(o_t\mid s_t)$, \textit{i.e.}, pixel-level reconstruction of observation $o_t$ from agent state $s_t$. The generative parameterization unavoidably models the entire observation space, including the complex but irrelevant information to decision making. For example in robot navigation, a generative model will try to capture the pixel-level distribution of the paintings on walls, which is irrelevant to the task of navigating to the goal. As a result, standard MBRL based on generative models have a more difficult optimization landscape given complex observations and will be ineffective when applied to natural environments.

In this paper, we present \textit{Contrastive Variational Reinforcement Learning} (CVRL) for robust MBRL under complex observations with high sample-efficiency and long-horizon planning ability. To be robust to complex observations, CVRL learns a \textit{contrastive variational world model} by discriminative contrastive learning, which captures the environment dynamics without modeling the complex observations. Specifically, CVRL maximizes the mutual information between state $s_t$ and observation $o_t$ by scoring the real pair $(s_t, o_t)$ against the fake pairs $\{(s_t, o')\}$ using a simple non-negative function. As a result, contrastive learning avoids directly modeling complex observations and is more robust than the generative models. For example, by contrasting observations from different places, the mini-cheetah can identify its current position $s_t$ by simply understanding what observations $\{o'\}$ are unlikely to receive. 
Mathematically, we derive a \textit{contrastive evidence lower bound} (CELBO), a new lower bound of $p(o_{1:T})$ from the mutual information perspective and it sidesteps the difficulty of learning a complex generative latent world model. CVRL solves the decision making problem combining online model predictive control (MPC)~\citep{camacho2013model} with learned heuristics, \textit{i.e.}, an efficiently and robustly trained actor-critic, for learning long-horizon behavior. 

We evaluate CVRL on three challenging continuous control domains: Mujoco tasks designed in Deepmind Control Suite~\citep{tassa2018deepmind}, Natural Mujoco tasks, a new domain with more complex observations that we introduce, and Box Pushing, a robot pushing experiment in PyBullet simulator~\citep{coumans2019}. CVRL outperforms the state-of-the-art model-based RL method in most cases. Results show that CVRL significantly improves the MBRL performance with contrastive representation learning. 

\begin{figure}[t]
	\centering
	\begin{tabular}{c c c c c}
	\includegraphics[width=0.14\linewidth]{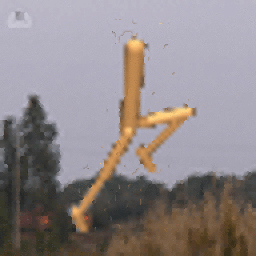} &
    \includegraphics[width=0.14\linewidth]{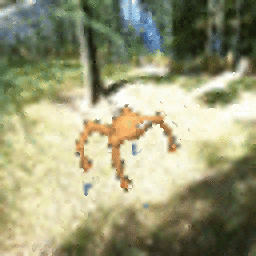} &
    \includegraphics[width=0.14\linewidth]{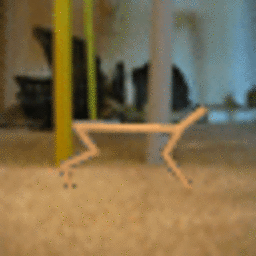} &
    \includegraphics[height=6em]{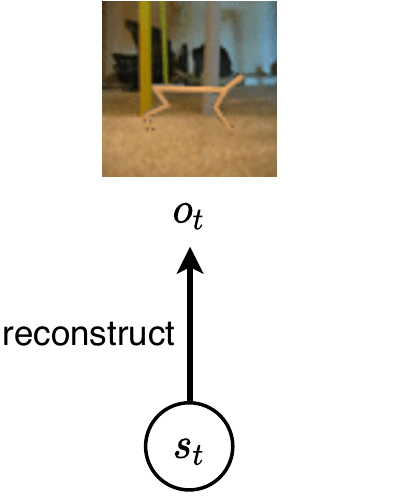} &
    \includegraphics[height=6em]{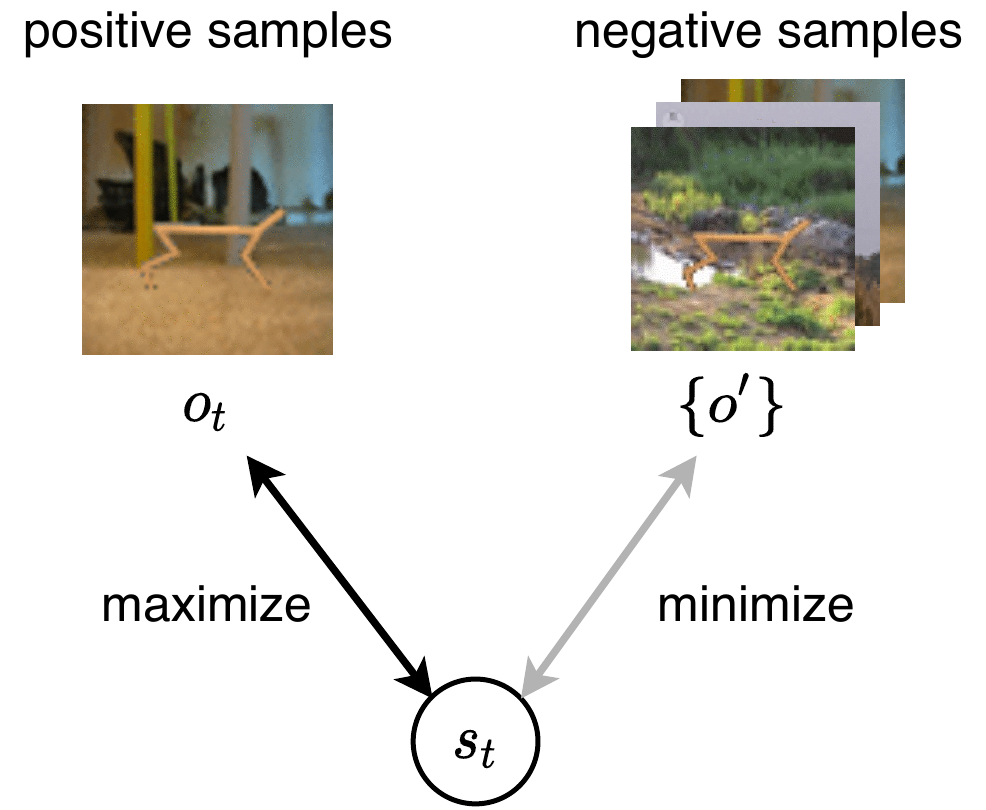} \\  
    \multicolumn{3}{c}{\small (a) Natural Mujoco Tasks} & \small (c) Generative Model & \small (b) Contrastive Model \\
	\end{tabular}
	\centering
	\caption{
	\small (a) CVRL addresses the tasks of sparse rewards, many degrees of freedom, and complex observations. (b) Standard generative observation model learns a observation likelihood function $p(o_t\mid s_t)$, \textit{i.e.}, reconstructing observations $o_t$ from $s_t$, which includes the irrelevant background features. (c) CVRL discovers the correspondence between state $s_t$ and observation $o_t$ by maximizing their mutual information $I(s_t, o_t)$ by scoring the real pair $(s_t, o_t)$ against fake pairs $\{s_t, o'_t\}$, which avoids pixel-level reconstruction.
	}
	\vspace{-5mm}
	\label{fig:mujoco}
\end{figure}

\section{Related Works}

\textbf{MBRL with World Models.}
Classic MBRL approaches have focused on planning in a predefined low-dimensional state space~\citep{doya2002multiple}. However, manually specifying a world model is difficult~\citep{karkus2017qmdp,ma2019particle}. Recently several works demonstrated that we could learn world models from raw pixel inputs. The majority rely on sequential variational autoencoders, which aims to minimize the reconstruction loss of the observations, to capture the stochastic dynamics of the environment~\citep{igl2018deep,hafner2018planet,hafner2019dream}. Some other works in robotics learn to predict videos directly for planning~\citep{agrawal2016learning,finn2017deep}.
However, real-world observations are complex and noisy, building an accurate generative model over the entire observation space is challenging, which leads to an accumulated compositional error of the world model. 

\textbf{Contrastive Learning.}
Contrastive learning are widely used for learning word embeddings~\citep{mnih2012fast}, image representation learning~\citep{pedagadi2013local}, graph representation learning~\citep{grover2016node2vec} and etc. The main idea is to construct real and fake sample pairs and use a function to score them in different ways. Concurrent to our work, contrastive learning has been applied to learn latent world models~\citep{hafner2019dream,kipf2019contrastive}, motivated from different perspectives. Specifically, Hafner et al.~\citep{hafner2019dream} use contrastive learning as an alternative to image reconstruction, where the contrastive learned agent gives worse performance compared with the one learned by image reconstruction. On the contrary, we would like to emphasize the strength of contrastive learning in handling complex visual observations. CVRL significantly outperforms the SOTA model~\citep{hafner2019dream} on tasks with complex observations.

\textbf{Reinforcement Learning under Complex Observations.}
Given complex observations, discriminative training is generally used to improve the robustness of the agent. Recent works suggest that learning task-oriented observation functions by end-to-end training improves the robustness of observation models~\citep{karkus2018particle,ma2019particle,ma2020discriminative,karkus2020differentiable}. In particular, Ma et al.~\citep{ma2020discriminative} introduced DPFRL which successfully addressed a challenging task with natural video in the background as well as robot navigation in a simulator constructed from real-world data. However, DPFRL relies on only the RL signal and is sample inefficient compared to model-based approaches. Besides, the generalization ability of DPFRL is also limited due to the model-free policy, and it failed on specific games. CVRL addresses the complex observation from a different perspective: we use contrastive learning to learn the latent world model, which avoids the modeling the complex observations. CVRL benefits both the sample efficiency of the model-based approaches and the robustness of the model-free approaches.

\begin{figure}[t]
	\centering
	\begin{tabular}{c  c}
    \includegraphics[width=0.45\linewidth]{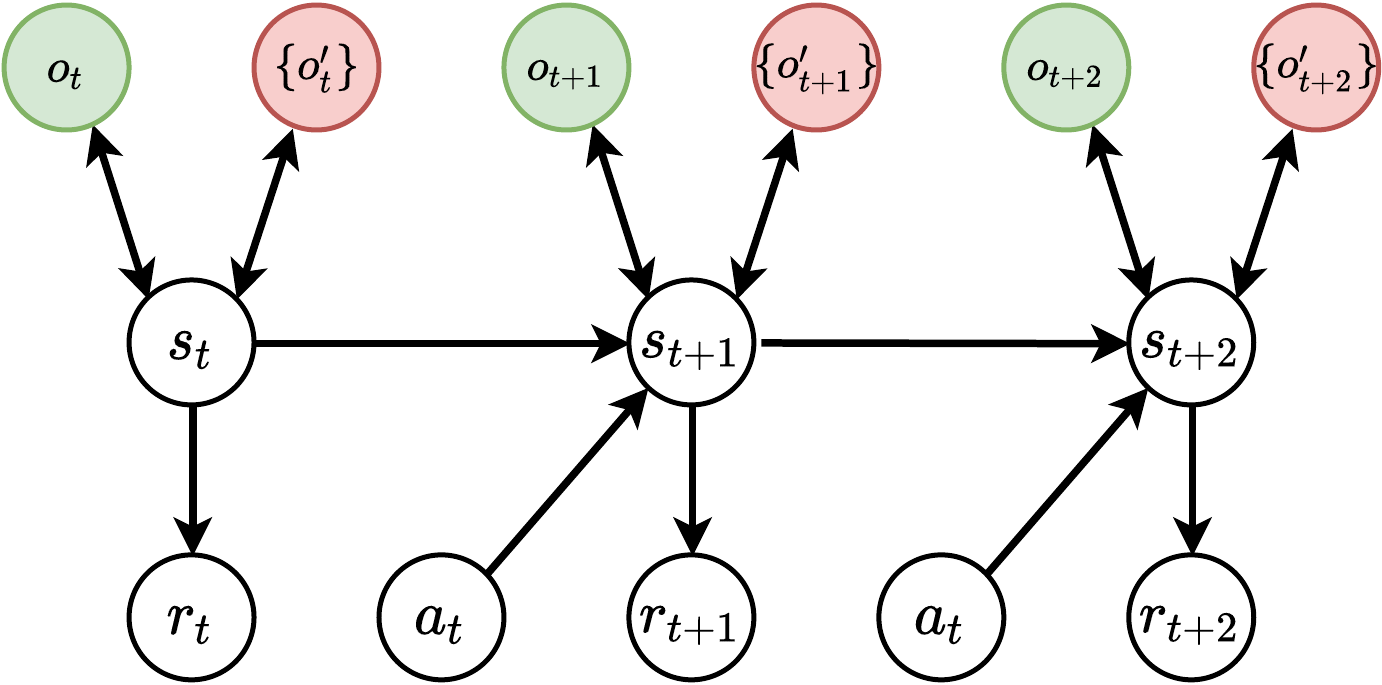} &
    \includegraphics[width=0.5\linewidth]{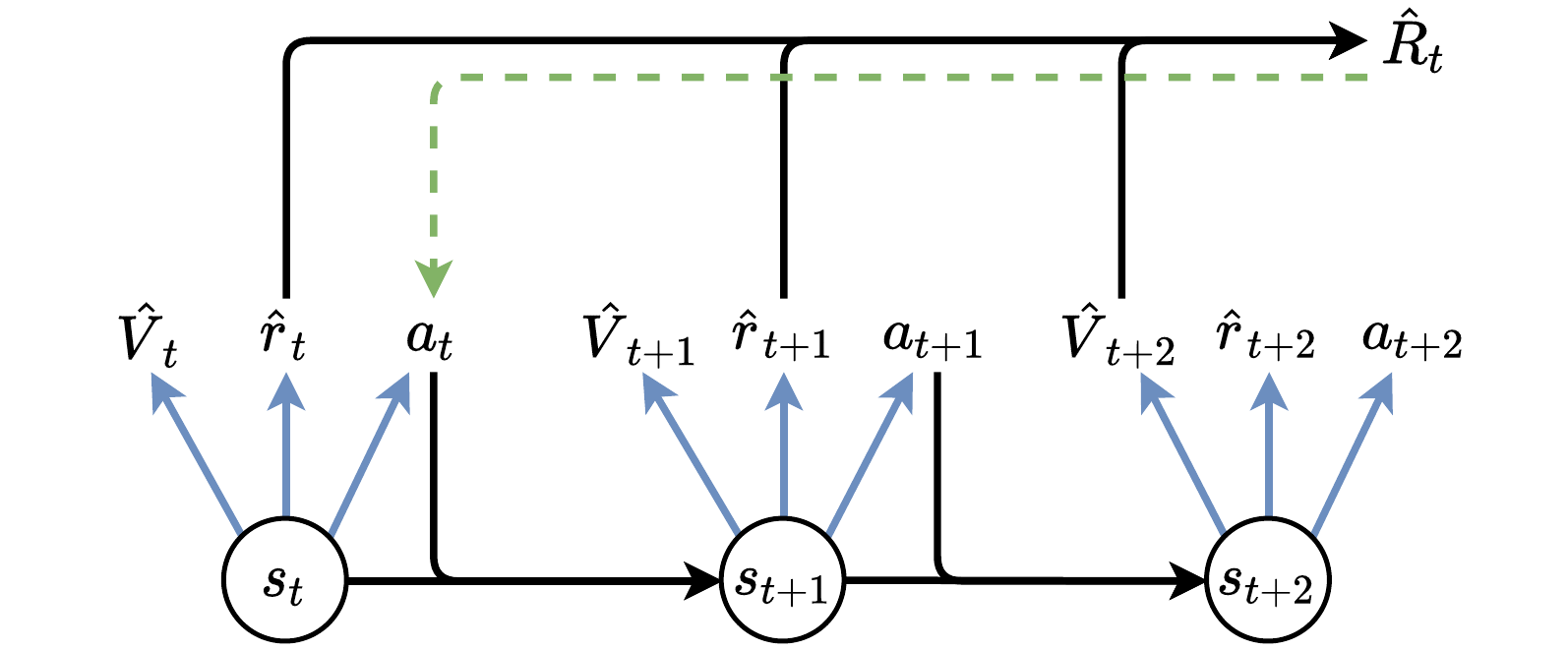}\\
    \small (a) Contrastive Latent World Model &\small (b) Latent Guided MPC
	\end{tabular}
	\centering
	\caption{\small (a) CVRL follows a contrastive latent world model, where the latent states are discovered by contrastive learning, \textit{i.e.}, maximizing the correspondence between state $s_t$ and the positive sample $o_t$ (in green) and minimizing the correspondence between a set of negative samples $\{o'_t\}$ (in red). 
	(b) CVRL chooses actions with a latent guided MPC using latent analytic gradients, which combines online planning with learned heuristics, \textit{i.e.}, an efficiently and robustly learned actor-critic.}
	\label{fig:CVRL}
\end{figure}

\section{Contrastive Variational Reinforcement Learning}
We introduce \textit{contrastive variational reinforcement learning} (CVRL) for reinforcement learnign with complex visual observations (Fig.~\ref{fig:CVRL}). 
In general, the visual observation reveals only part of the true underlying state. We thus treat the visual control task as a partially observable Markov decision process (POMDP) with discrete time  $t=1,2, \ldots, T$, continuous action $a_t$,  visual observation $o_t$, and scalar reward $r_t$. CVRL learns a \textit{contrastive latent world model}, which consists of a probabilistic transition function $p(s_t\mid s_{t-1}, a_t)$, a reward function $p(r_t\mid s_t)$, and a ``discriminative" observation function $f(o_t, s_t)$ through contrastive learning. Contrastive learning scores a positive state-observation pair $(s_t, o_t)$ against a set of negative observations $\{o'_t\}$ by maximizing $f(s_t, o_t)$ while minimizing $f(s_t, o'_t)$. Contrastive learning is significantly more robust than generative learning, as it avoids pixel-level reconstruction of complex observations. We introduce a new optimization objective, \textit{Contrastive Evidence Lower Bound} which provides a  lower bound on the generative optimization objective.  CVRL performs decision making by \textit{Latent Guided Model Predictive Control} (MPC) using analytic gradients with learned dynamics. The latent guided MPC combines online planning with learned heuristics, \textit{i.e.}, an efficiently and robustly learned \textit{Hybrid Actor-Critic} model.

\subsection{Variational Latent World Models}\label{sect:rssm}
CVRL adopts a variational latent model for discovering environment dynamics from pixel inputs. Variational latent world models are the sequential version of variational autoencoders (VAEs)~\citep{kingma2013auto}. For an observable variable $x$, VAEs learn a latent variable $z$ that generates $x$ by optimizing an \textit{Evidence Lower Bound} (ELBO) of $\log p(x)$
\begin{equation}
    \log p(x) = \log \int_z p(x\mid z)p(z)dz \geq \mathbb{E}_{q(z\mid x)}\left[ p(x\mid z) \right] - KL\left[ q(z\mid x)\parallel p(z) \right]
\end{equation}
where $KL[q(z\mid x)\parallel p(z)]$ denotes the Kullback–Leibler divergence between the prior distribution $p(z)$ and a proposal distribution $q(z\mid x)$ that samples $z$ conditioned on $x$.

In a sequential decision making task, CVRL applies a multi-step generalization to the single step ELBO by optimizing an ELBO of $p(o_{1:T}, r_{1:T}\mid a_{1:T})$~\citep{igl2018deep,hafner2018planet}
\begin{align}
&\log p(o_{1:T}, r_{1:T}\mid a_{1:T}) = \log\int p_\theta(s_t\mid s_{t-1}, a_{t-1})p_\theta(o_t\mid s_t)p_\theta(r_t\mid s_t)ds_{1:T}\nonumber\\
& \geq \sum\limits_{t=1}^T \left( 
\underbrace{\underset{q_\phi(s_t\mid o_{\leq t}, a_{\leq t})}{\mathbb{E}\left[ \log p_\theta(o_t\mid s_t) \right]} + \underset{q_\phi(s_t\mid o_{\leq t}, a_{\leq t})}{\mathbb{E}\left[\log p_\theta(r_t\mid s_t) \right]}}_{\mbox{reconstruction}}
- 
\underbrace{\underset{q_\phi(s_{t-1\mid o_{\leq t-1}, a_{<t-1}})\qquad \qquad\qquad\qquad\qquad}{\mathbb{E}\left[ KL[q_\phi(s_t\mid o_{\leq t}, a_{\leq t})] \parallel p_\theta(s_t\mid s_{t-1}, a_{t-1}) \right]}}_{\mbox{dynamics}}  
\right)\label{eqn:elbo}
\end{align}
where $\theta$ and $\phi$ are model parameters. The first part encourages accurate reconstructions of the observation likelihood $p_\theta(o_t\mid s_t)$ and reward likelihood $p_\theta(r_t\mid s_t)$; the second part encourages learning self-consistent dynamics by KL-divergence. Specifically, the second part minimizes the KL divergence between the prior distribution $p_\theta(s_t\mid s_{t-1}, a_{t-1})$ and the posterior distribution $q_\phi(s_t\mid o_{\leq t}, a_{<t})$ conditioned on the observation sequences.

However, the pure stochastic transitions might have difficulties remembering the history and learning stability. Introducing a sequence of additional deterministic states $h_{1:T}$ tackles this issue~\citep{chung2015recurrent,hafner2018planet}. In this work, we use the recurrent state space model (RSSM)~\citep{hafner2018planet} that decomposes the original latent dynamic model into the following four components

\begin{tabular}{ll}
	Deterministic state model: $h_t = f_\theta(h_{t-1}, s_{t-1}, a_{t-1})$& Stochastic state model: $s_t\sim p_\theta(s_t\mid h_t)$\\   
	Observation model: $o_t\sim p_\theta(o_t\mid s_t)$ & Reward model: $r_t\sim p_\theta(r_t\mid h_t, s_t)$ 
\end{tabular}

As a result, during training, RSSM approximate $q_\phi(s_t\mid o_{\leq t}, a_{\leq t})$ by $q_\phi (s_t\mid h_t, o_t)$.

\subsection{Contrastive Evidence Lower Bound}

One big issue of RSSM is that the pixel level generative observation model $p(o_t\mid s_t)$ has to capture the entire observation space, which is problematic given complex observations, \textit{e.g.}, natural observations in autonomous driving or the natural Mujoco games. Given various videos, pixel-level reconstruction becomes difficult which leads to the inaccuracy in the learned latent world model. We introduce \textit{Contrastive Evidence Lower Bound} (CELBO), a robust optimization objective that avoids reconstructing the observations and lower bounds the original ELBO (Eqn.~\ref{eqn:elbo}). 

Instead of maximizing the observation likelihood $p(x\mid z)$, we motivate the solution from a mutual information perspective. The mutual information between two variables $x$ and $y$ is defined as
\begin{equation}
    I(x, y) = \int p(x, y) \log \frac{p(x,y)}{p(x)p(y)} dxdy = \mathbb{E}_{p(x,y)}\left[ \log \frac{p(x\mid y)}{p(x)} \right]
\end{equation}
In Eqn.~\ref{eqn:elbo}, the observation likelihood is computed for a specific trajectory $\tau = \{o_{1:T}, r_{1:T}, a_{1:T}\}$. In practice, during optimization, we consider the observation likelihood over a distribution of $\tau$. We can rewrite the observation likelihood in Eqn.~\ref{eqn:elbo} as
\begin{equation}
    \underset{q_\phi(s_t\mid o_{\leq t}, a_{\leq t})p(o_{\leq t}, a_{\leq t})\qquad\qquad\qquad\quad}{\mathbb{E}\left[ \log p_\theta(o_t\mid s_t) - \log p(o_t) + \log p(o_t) \right]} = I(s_{t}, o_{t}) + E_{p(o\leq t)}\left[ \log p(o_t) \right]\label{eqn:mutual_info}
\end{equation}
where the second term $E_{p(o\leq t)}\left[ \log p(o_t) \right]$ could be treated as a constant that can be ignored during optimization. Eqn.~\ref{eqn:mutual_info} suggests that maximizing the observation likelihood is equivalent to maximizing the mutual information of the state-observation pairs. The benefit of such a formulation is that mutual information could be estimated without reconstructing the observations, \textit{e.g.}, using energy models~\citep{lecun2006tutorial} or the ``compatibility function"~\citep{karkus2018particle,ma2020discriminative}. 
When the observations are complex, mutual information formulation is more robust than the generative parameterization.

To efficiently optimize the mutual information, we use the InfoNCE, which is a contrastive learning method that optimizes a lower bound of the mutual information~\citep{oord2018representation} and is proven to be powerful in a set of self-supervised learning tasks~\citep{oord2018representation,velickovic2019deep}. Using the result in InfoNCE, the mutual information $I(s_t, o_t)$ could be lower bounded by 
\begin{equation}
    I(s_{t}, o_{t}) \geq \mathbb{E}_{q_\phi(s_t\mid o_{\leq t}, a_{\leq t})p(o_{\leq t}, a_{\leq t})}\left[\log f_\theta(s_t, o_t) - \log \sum_{o_t'\in \mathcal{O}_t} f_\theta(s_t, o_t') \right]\label{eqn:info_nce}
\end{equation}
where function $f_\theta (s_t, o_t)$ is a non-negative function that measures the compatibility between state $s_t$ and observation $o_t$, and $\mathcal{O}_t$ is a set of irrelevant observations sampled from a replay buffer. An intuition for Eqn.~\ref{eqn:info_nce} is that we want to maximize the compatibility between the state $s_t$ and the real observation $o_t$ (positive sample), while minimizing its compatibility between a set of irrelevant observations (negative samples). In our case, we follow the setup of the original InfoNCE loss and use a simple bi-linear model for $f_\theta (s_t, o_t) = \exp (z_t^\mathrm{T}W_\theta s_t)$, where $z_t$ is an embedding vector for observation $o_t$ and $W_\theta$ is a learnable weight matrix parameterized by $\theta$.

Substituting Eqn.~\ref{eqn:mutual_info} and Eqn.~\ref{eqn:info_nce} into Eqn.~\ref{eqn:elbo}, we derive the CELBO of $p(o_{1:T}, r_{1:T}\mid a_{1:T})$ as
\begin{align}
&\log p(o_{1:T}, r_{1:T}\mid a_{1:T})\geq\nonumber\\
    &\sum\limits_{t=1}^T \left( 
    \underbrace{\underset{q_\phi(s_t\mid o_{\leq t}, a_{\leq t})\qquad\qquad}{\mathbb{E}\left[ \log \frac{f_\theta(o_t, s_t)}{\sum_{o_t'\in \mathcal{O}_t} f_\theta (s_t, o_t')} \right]}}_{\mbox{contrastive learning}} 
    + 
    \underbrace{\underset{q_\phi(s_t\mid o_{\leq t}, a_{\leq t})}{\mathbb{E}\left[\log p_\theta(r_t\mid s_t) \right]}}_{\mbox{reconstruction}}
    - 
    \underbrace{\underset{q_\phi(s_{t-1\mid o_{\leq t-1}, a_{<t-1}})\qquad \qquad\qquad\qquad\qquad}{\mathbb{E}\left[ KL[q_\phi(s_t\mid o_{\leq t}, a_{\leq t})] \parallel p_\theta(s_t\mid s_{t-1}, a_{t-1}) \right]}}_{\mbox{dynamics}}
    \right)\label{eqn:celbo}
\end{align}

The CELBO objective is similar to the Deep Variational Information Bottleneck~\citep{alemi2016deep} in the sense of mutual information maximization. The difference is that we take a mixed approach: we use contrastive learning to optimize the mutual information for only the state-observation pairs, and maximize the reward likelihood $p(r_t\mid s_t)$. Compared to the complex observations, the scalar reward is easy to reconstruct. The quality of contrastive learning highly depends on the choice of negative samples. Reward reconstruction is easier to optimize compared to contrastive learning. 

\subsection{Hybrid Actor-Critic}
CVRL trains an actor-critic using a hybrid-approach, benefiting from the sample-efficiency of the model-based learning and the task-oriented feature learning from the model-free RL. 

\textbf{Actor-Critic from Latent Imagination.}
First, CVRL uses latent imagination to train the actor-critic, 
\textit{i.e.}, reasoning the latent world model, which reduces the amount of the interactions needed with the non-differentiable environment. 
In particular, since the predicted reward and latent dynamics are differentiable, the analytic gradients can back-propagate through the dynamics. As a result, the actor-critic can potentially approximate long-horizon planning behaviors~\citep{hafner2019dream}.

We adopt the same strategy with Dreamer~\citep{hafner2019dream}. We parameterize the actor model $a_t\sim q_\eta (a_t\mid s_t)$ as a tanh-transformed Gaussian, \textit{i.e.}, $a_t = \mathrm{tanh}(\mu_\eta(s_t) + \sigma_\eta (s_t)\epsilon)$, where $\epsilon\sim \mathcal{N}(0, \mathbb{I})$. For value model, we use a feed-forward network $v_\psi (s_t)$ with a scalar output. To compute the analytic gradient, we first estimate the state values of the imagined trajectory $\{\Tilde{s}_\tau, \Tilde{a}_\tau, \Tilde{r}_\tau\}_{\tau=t}^{t+H}$, where the actions are sampled from the actor network. We denote the value estimate of $s_\tau$ as a function $\Tilde{V}(s_\tau)$. Detailed descriptions of the value estimation and imagined trajectory generation are in the appendix. The Dreamer learning objective is thus given by
\begin{equation}
    L_{\mathrm{Dreamer}} = \underbrace{-\mathbb{E}_{p_\theta, q_\eta}\left[ \sum\limits_{\tau=t}^{t+H} \Tilde{V}(s_\tau) \right]}_{\mbox{actor loss}}  + \underbrace{\mathbb{E}_{p_\theta, q_\eta}\left[ \sum\limits_{\tau=t}^{t+H}\frac{1}{2} \parallel v_\psi (s_\tau) - \Tilde{V}(s_\tau) \parallel^2 \right]}_{\mbox{critic loss}}
    \label{eqn:dreamer}
\end{equation}

\textbf{Hybrid Actor-Critic.}
The performance of latent imagination highly relies on the accuracy of the learned latent world model. Given complex observations, learning an accurate world model is difficult, even with CELBO. We introduce a simple yet effective hybrid training scheme to address this issue. CVRL combines the Dreamer objective with a secondary training signal from off-policy RL, using the ground truth trajectories. Discriminative RL objective can improve the robustness of the actor-critic, while sacrificing the sample-efficiency~\citep{ma2020discriminative}. Thus, CVRL benefits from both the sample-efficiency of the latent analytic gradient and the robustness of discriminative RL gradient.

In our experiment, we use the Soft Actor-Critic (SAC)~\cite{haarnoja2018soft} to perform off-policy RL. During each optimization step, we use the ground truth trajectory $\{s_t, a_t, r_t\}_{t=1}^T$, and use the imagined trajectories $\{\Tilde{s}_\tau, \Tilde{a}_\tau, \Tilde{r}_\tau\}_{\tau=t}^{t+H}$. We have the final objective as $L_\mathrm{CVRL} = L_\mathrm{Dreamer} + \alpha * L_\mathrm{SAC}$.

\subsection{Latent Guided Model Predictive Control}
Although the learned actor-critic maximizes the accumulated rewards, a model-free policy, without explicit reasoning with world models, might be stuck in local optimum~\citep{karkus2017qmdp,karkus2019dan}. Model predictive control (MPC) is widely used to address the continuous control problems, where multiple iterations are required for the policy to converge to the optimal solution~\citep{tedrake2009underactuated}.

We introduce latent guided model predictive control. Specifically, we use the shooting method in trajectory optimization to address the MPC task. For state $s_t$, we perform a forward search using the latent world model guided by the learned actor-critic, and generate latent imagination trajectory $\{\Tilde{s}_\tau, \Tilde{a}_\tau, \Tilde{r}_\tau\}_{\tau=t}^{t+H}$.  We compute the value estimate for the sampled trajectory using $\Tilde{V}(s_t)$, compute the analytic gradient by maximizing $\Tilde{V}(s_t)$ and update the action sequences with analytic gradients. In practice, the combination of the offline training with online planning gives a better performance. The detailed description of the algorithm can be found in the appendix.

\section{Experiments}

We first evaluate CVRL on 10 continuous Mujoco control tasks in Deepmind Control Suite~\citep{tassa2018deepmind}. We then introduce a new, more challenging benchmark, Natural Mujoco. Finally, we apply CVRL to a robot pushing task with RGB images and randomized lighting sources in PyBullet simulator~\citep{coumans2019}.

We compare CVRL with the SOTA generative MBRL method, Dreamer~\citep{hafner2019dream}, and a model-free baseline, Soft Actor-Critic~\citep{haarnoja2018soft}\footnote{we use the official implementation of Dreamer and the SAC implementation from OpenAI baselines}. We also include the result of D4PG~\citep{barth2018distributed} trained for sufficient time on standard Mujoco tasks, as a baseline for Mujoco tasks. All results are averaged over 3 random seeds. A detailed description to the experiment setup can be found in the appendix. We show that: 1) CVRL significantly outperforms SAC in all cases, with much fewer training iterations; 2) CVRL significantly outperforms Dreamer on natural Mujoco tasks because of the robust contrastive learning, and achieves comparable performance on standard Mujoco; 3) the proposed hybrid actor-critic training scheme and guided model predictive control further improves the performance of CVRL.

\subsection{Mujoco Tasks}

\begin{table*}[t]
\centering

\fontsize{8}{8}\selectfont
\begin{tabular}{lccccccc}
\toprule
\multirow{2}{*}{} & \multicolumn{4}{c}{Standard}     & \multicolumn{3}{c}{Natural} \\
\cmidrule(lr){2-5}\cmidrule(lr){6-8}
                  & CVRL    & Dreamer$^\dagger$~\citep{hafner2019dream}   & SAC & D4PG$^\dagger$~\citep{hafner2019dream}   & CVRL    & Dreamer   & SAC   \\
\midrule
walker-walk                           & \textbf{980.3} & 961.7          & 355.7 & 968.3          & \textbf{941.5} & 206.6   & 44.1  \\
walker-run                            & 377.7          & \textbf{824.6} & 153.0 & 567.2          & \textbf{382.1} & 82.7    & 78.1  \\
cheetah-run                           & 528.1          & \textbf{894.5} & 181.8 & 523.8          & \textbf{248.7} & 100.7   & 35.8  \\
finger-spin                           & \textbf{717.8} & 498.8          & 258.5 & 985.7          & \textbf{850.4} & 13.6    & 23.8  \\
cartpole-balance                      & \textbf{997.1} & 979.6          & 355.5 & 992.8          & \textbf{911.9} & 163.7   & 206.0 \\
catpole-swingup                       & \textbf{863.4} & 833.6          & 252.5 & 862.0          & \textbf{413.8} & 117.6   & 150.5 \\
cup-catch                             & \textbf{964.9}          & 962.5          & 421.4 & 980.5 & \textbf{894.2} & 131.1   & 202.2 \\
reacher-easy                          & \textbf{968.2} & 935.1          & 239.2 & 967.4          & \textbf{909.1} & 133.7   & 137.7 \\
quadruped-walk                        & \textbf{950.3} & 931.6          & 337.1 & -              & \textbf{878.7} & 153.2   & 204.3 \\
pendulum-swingup                      & \textbf{912.1} & 833.0          & 28.6  & 680.9          & \textbf{842.9} & 12.4    & 14.8         \\

\bottomrule
\end{tabular}
\caption{\small CVRL achieves comparable performance with the SOTA method, Dreamer~\citep{hafner2019dream}, on standard Mujoco tasks and significantly outperforms Dreamer on Natural Mujoco tasks. CVRL, Dreamer and SAC are trained for $5\times 10^6$ steps, while the best model-free baseline D4PG is trained for $1\times 10^8$ steps, which we use as an indicator for the performance in standard Mujoco tasks.
$^\dagger$Results are taken directly from Dreamer paper.}
\label{tab:results}
\end{table*}
\begin{figure}[t]
	\centering
	\begin{tabular}{c c c c c}
    \includegraphics[width=0.13\linewidth]{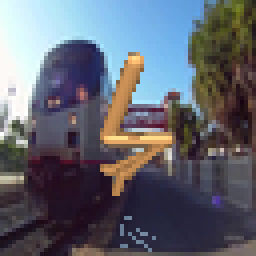} & \includegraphics[width=0.13\linewidth]{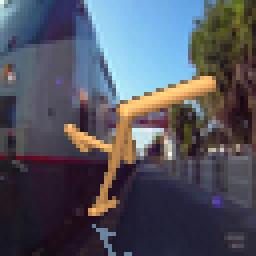} & \includegraphics[width=0.13\linewidth]{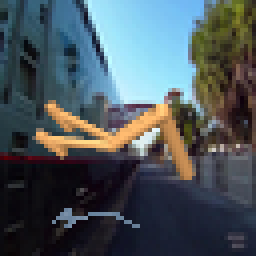} & \includegraphics[width=0.13\linewidth]{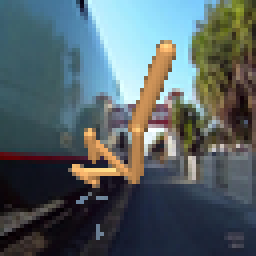} & \includegraphics[width=0.13\linewidth]{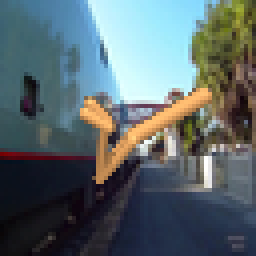} \\
    \includegraphics[width=0.13\linewidth]{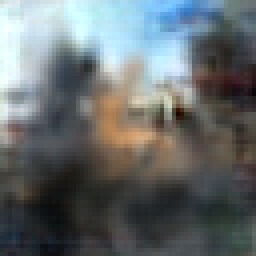} & \includegraphics[width=0.13\linewidth]{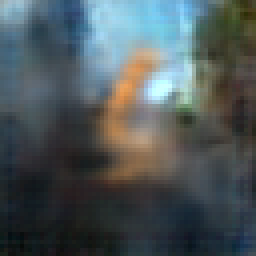} & \includegraphics[width=0.13\linewidth]{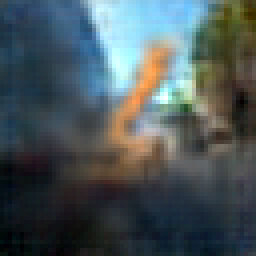} & \includegraphics[width=0.13\linewidth]{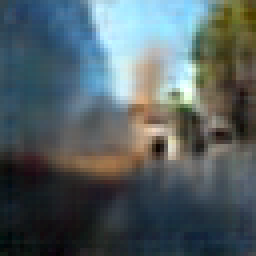} & \includegraphics[width=0.13\linewidth]{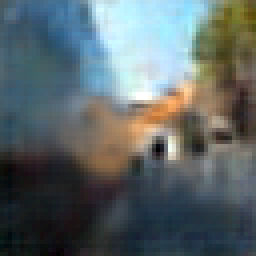} \\
    \small $t=1$ & \small $t=10$ & \small $t=20$ & \small $t=30$ & \small $t=40$
	\end{tabular}
	\centering
	\caption{\small Generative models learn a latent world model by pixel level reconstruction, which is difficult when the observations are complex and variable. The first row shows the complex observations of natural Walker with varying video backgrounds, and the second row shows the reconstruction of generative models. }
	\vspace{-5mm}
	\label{fig:reconstructions}
\end{figure}

The Mujoco tasks are difficult for reinforcement learning methods due to the sparse reward, 3D scenes and contact dynamics. Specifically, Natural Mujoco tasks are significantly more challenging, where they bridge the gap between simulated environment and real robots by replacing the simple backgrounds with natural videos sampled from ILSVRC dataset~\citep{russakovsky2015imagenet}. We present the results in Table~\ref{tab:results} and analyze the quantitative results as follows.

Specifically, Natural Mujoco tasks brides the gap between simulated environment and real robots by replacing the simple backgrounds with natural videos sampled from ILSVRC dataset~\citep{russakovsky2015imagenet}.

\textit{Model-based CVRL outperforms the model-free baseline.} We observe that both CVRL reaches the best achievable performance, indicated by D4PG, the state-of-the-art model-free baseline trained for 20 times more steps ($5\times 10^6$ steps for CVRL and Dreamer, and $1\times 10^8$ steps for D4PG). The learned latent world model successfully captures the real environment dynamics from pixel-level input, so that the trained actor-critic achieves comparable performance with the SOTA D4PG trained by ground truth trajectories. In contrast, given the same number of training steps, CVRL and Dreamer significantly outperform SAC on all tasks. This also suggests that the benefit of CVRL comes from the overall framework design, rather than the SAC.

\textit{CVRL is more robust to the natural observations.} In Natural Mujoco tasks where the observations are more complex and variable, CVRL significantly outperforms the generative Dreamer in all cases. Although Dreamer achieves SOTA performance on the standard Mujoco tasks with relatively simple observations, its performance drops dramatically on natural Mujoco given complex observations introduced by the video background (\textit{e.g.}, on walker-walk, 961.7 V.S. 206.6). CVRL, however, achieves comparable performance on 8 out of 10 tasks with or without the video background (\textit{e.g.}, on walker-walk, 980.3 V.S. 941.5). This suggests that the contrastive learning, which avoids the pixel-level reconstruction, helps to learn a more robust latent world model than the generative models. Even with the variable complex video background, the learned latent world still successfully captures the underlying dynamics and achieves comparable performance with the simple observations. Besides, we visualize the reconstruction of generative models in Fig.~\ref{fig:reconstructions}. 
The reconstructions are blurry and lose information about the agent, which explains the failure of the generative Dreamer.

\begin{figure}[!htb]
	\centering
	\begin{tabular}{ccc}
    \includegraphics[width=0.3\linewidth]{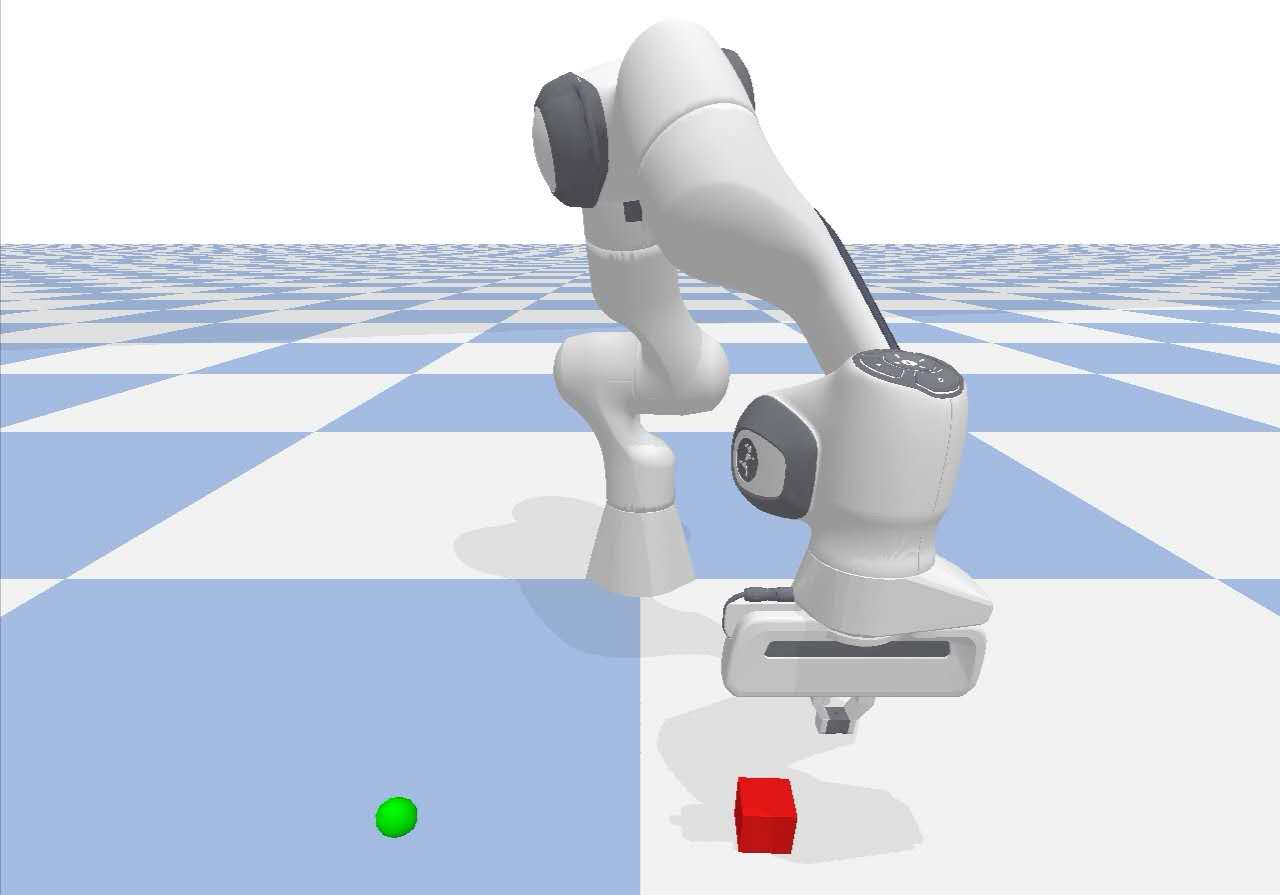} &
    \includegraphics[width=0.27\linewidth]{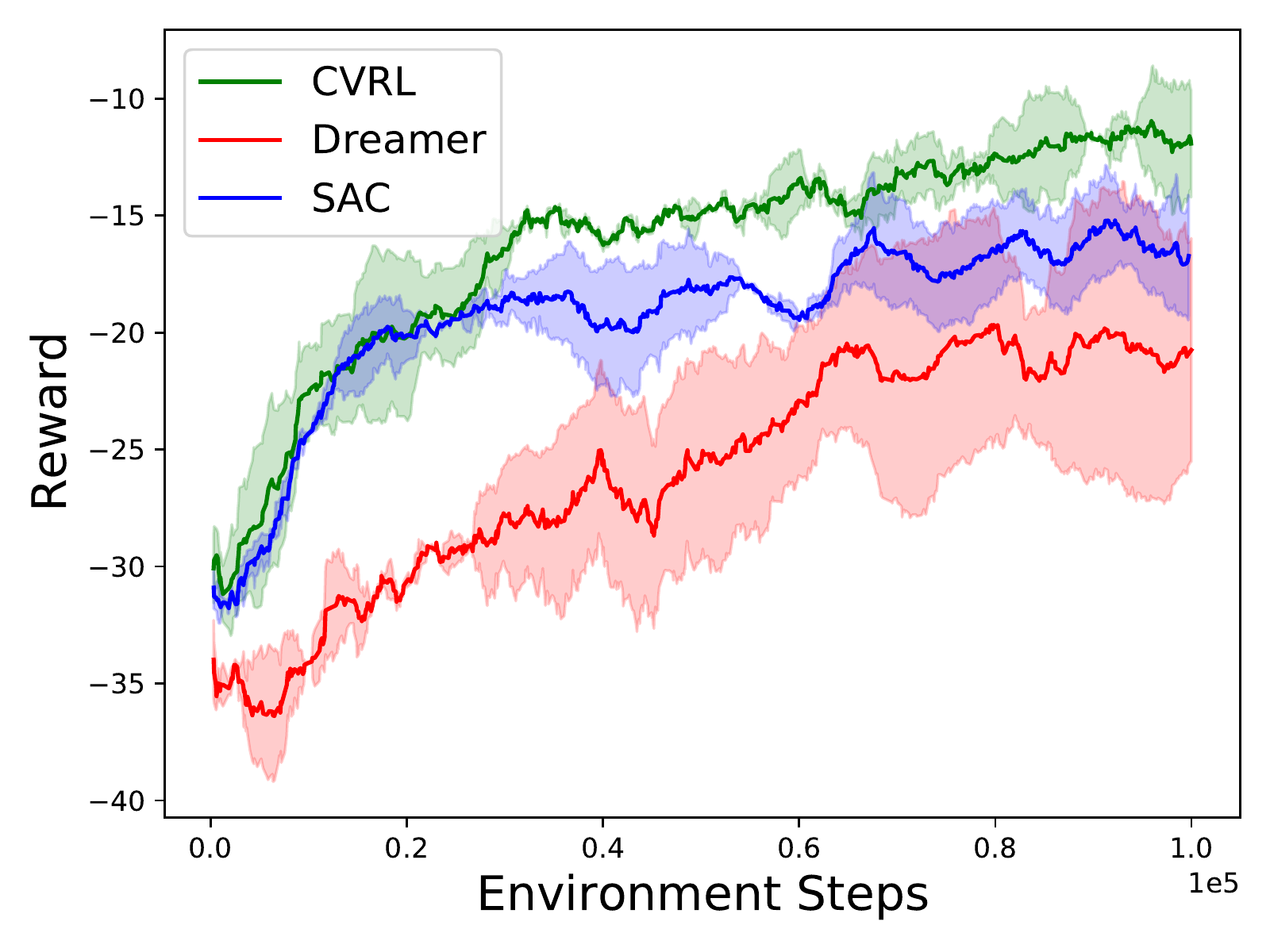} &
    \includegraphics[width=0.27\linewidth]{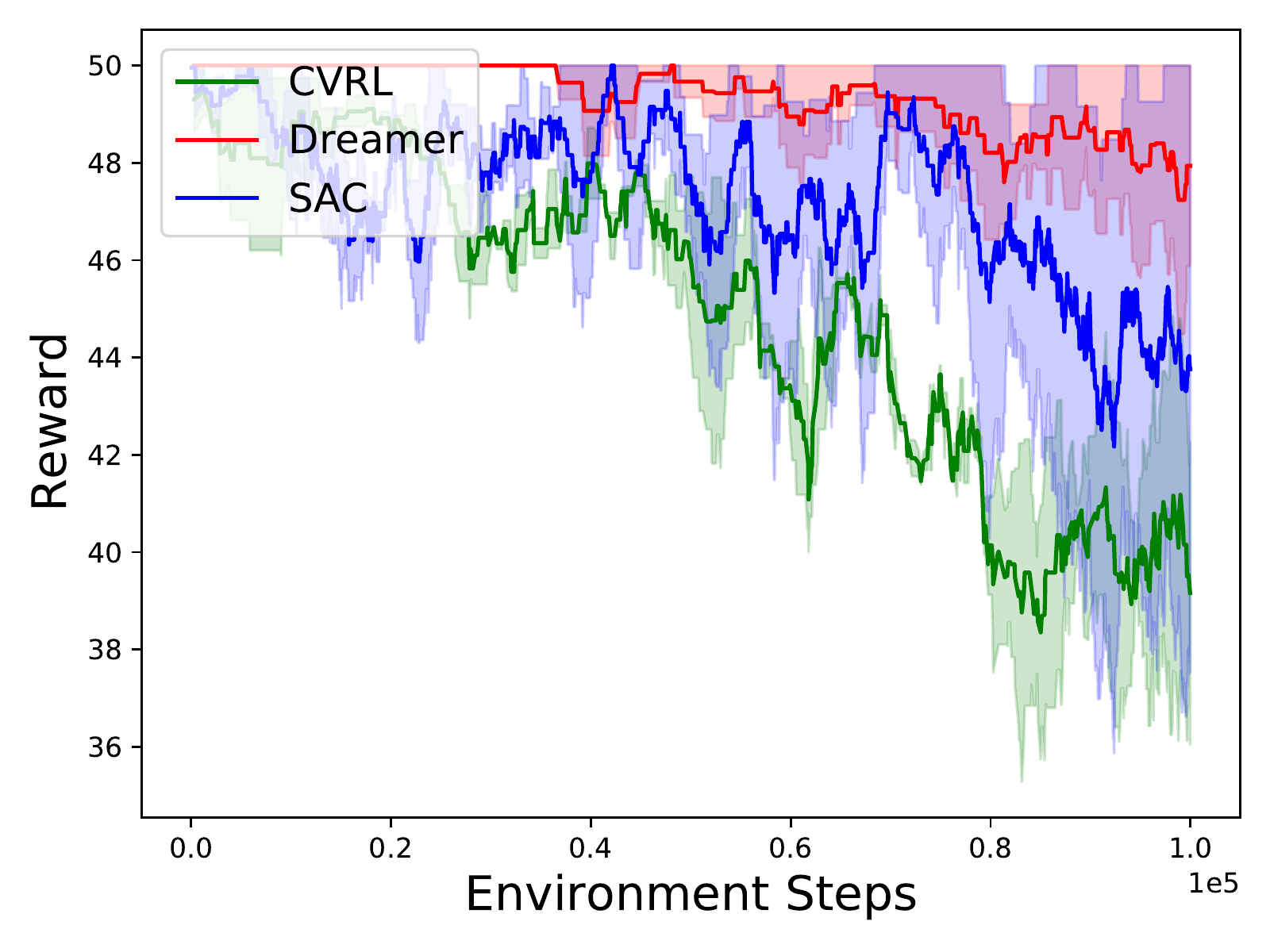}\\
    \small (a) Box Pushing Environment &\small (b) Accumulated Reward &\small (c) Execution Length
	\end{tabular}
	\centering
	\caption{\small Box Pushing environment is challenging due to the contact physics, 3D scene, and the changing shadow introduced by the robot arm, which commonly exists in robot manipulation scenarios. CVRL achieves the highest performance with the shortest execution length.}
	\label{fig:robot}
\end{figure}

\subsection{Box Pushing}
Robot pushing poses great challenges to reinforcement learning~\citep{li2018push}. Specifically, the shadow and the occlusion by the robot arm often introduce confusing information to the perception module. We evaluate CVRL for robot pushing in PyBullet which gives an efficient and high-quality physics simulation. The task is to push the red box to the goal indicated by a green ball with an action of $a = (dx, dy)$ for the positional displacement. The positions of the box, the goal, and the lightning source are randomized at every episode. The negative Euclidean distance from the object to the goal is used as the reward, and the robot receives a reward of +1 when the box reaches the goal.

The results are presented in Fig.~\ref{fig:robot}. We observe that CVRL achieves the highest reward with the shortest execution length, and it learns faster than Dreamer and SAC. Specifically, Dreamer struggles because reconstructing the shadow and the moving robot arm is challenging for generative learning. In contrast, although model-free SAC has failed on most of the Mujoco tasks with complex control dynamics, it outperforms Dreamer on Box Pushing by avoiding generative modeling, given the relatively simple action space. CVRL benefits from the sample efficiency of model-based learning, and it maintains the robustness to complex observations with contrastive learning.

\subsection{Ablation Studies}
\begin{table}[t]
\centering
\fontsize{8}{8}\selectfont
\begin{tabular}{lccccc}
\toprule
                    & CVRL           & CVRL-generative & CVRL-no-MPC     & CVRL-no-SAC    & CVRL-reward-only \\
\midrule
walker-walk          & \textbf{941.5} & 297.7           & 904.8          & 915.2          & 197.9            \\
walker-run           & \textbf{382.1} & 71.4            & 343.2          & 378.3          & 115.4            \\
cheetah-run          & 248.7          & 113.3           & \textbf{430.1} & 301.0          & 284.8            \\
finger-spin          & \textbf{850.4} & 13.9            & 753.3          & 668.8          & 68.7             \\
cartpole-balance     & 911.9          & 188.4           & \textbf{996.3} & 962.3          & 431.6            \\
catpole-swingup      & \textbf{413.8} & 160.5           & 353.0          & 465.9          & 176.3            \\
ball\_in\_cup-catch  & 894.2          & 254.8           & 881.4          & \textbf{930.4} & 368.7            \\
reacher-easy         & \textbf{909.1} & 235.8           & 858.9          & 880.5          & 167.2            \\
quadruped-walk       & \textbf{878.7} & 157.3           & 595.2          & 213.5          & 188.7            \\
pendulum-swingup     & \textbf{842.9} & 19.7            & 831.5          & 813.3          & 20.8                 \\
\bottomrule
\end{tabular}
\caption{\small Ablation Studies on natural Mujoco tasks. CVRL generally outperforms all other variants. 
}
\label{tab:ablations}
\vspace{-5mm}
\end{table}

We conduct a comprehensive ablation study on the Natural Mujoco tasks to better understand each proposed component. The results are presented in Table~\ref{tab:ablations}. 

\textit{Contrastive variational latent world model is more robust to complex observations.} CVRL-generative replaces the contrastive learning with a generative model that performs image-level reconstruction. Unlike Dreamer, CVRL-generative only differs from the CVRL in the parameterization of the representation learning method, and still has the rest of the proposed components. However, its performance degrades on all cases compared to CVRL. This aligns with our previous observation that contrastive learning is more robust given complex observations.

\textit{Latent guided MPC improves the reasoning ability for long-horizon behaviors.} 
CVRL-no-MPC uses only the actor-critic for decision making. We observe it performs worse than CVRL on 8 out of 10 tasks, especially on some of the challenging tasks, \textit{e.g.}, cartpole-swingup and quadruped-walk, where multi-step reasoning is required. The latent guided MPC improves the overall performance of CVRL.

\textit{The hybrid actor-critic is robust given complex observations.} CVRL-no-SAC removes the SAC during actor-critic learning. Its performance drops on certain cases, compared to CVRL (on cheetah-run, 497.3 V.S. 301.0 and finger-spin, 987.1 V.S. 668.8). This is because when the useful features are highly coupled with variable and complex background, learning an accurate latent world model becomes difficult, even for CELBO. With ground-truth trajectories, SAC can provide accurate training signals to compensate for the compositional error of the latent world model.

\textit{Reward signal alone is not enough for learning the latent world model.} CVRL-reward-only uses only reward prediction for representation learning. Its performance drops in all cases. This suggests that the robustness of CVRL comes from the contrastive learning, rather than only the reward learning.

\section{Conclusions}
We introduce CVRL, a framework for robust MBRL under natural complex observations.
CVRL learns a contrastive variational world model with CELBO objective, a contrastive learning alternative to the ELBO, which avoids reconstructing the complex observations. CVRL lerans a robust hybrid actor-critic and uses guided MPC for decision making. It achieves comparable performance with the SOTA methods on 10 challenging Mujoco control tasks, and significantly outperforms SOTA methods on more challenging Natural Mujoco tasks and Box Pushing tasks.

However, CVRL does not perform as well as Dreamer on some tasks on standard Mujcoo tasks (walker-run and cheetah-run), where the observation is simple. While contrastive learning is robust to complex observations, its quality highly depends on the sampling strategy of negative samples. Currently we use a very simple strategy. Further work may consider smarter sampling strategies, \textit{e.g.}, learning to sample using meta-learning.

\clearpage
\appendix

\section{Algorithm Details}

\subsection{Latent Imagination}\label{sect:latetn_imag}
CVRL first generates the imagined trajectories using the learned world model parameterized by $\theta$. Specifically, given a state $\ts_{\tau-1}$, we sample the next imagined state by $\ts_\tau\sim p_\theta (\ts_\tau\mid \ts_{\tau-1}, \ta_{\tau-1})$, which further generates a reward $\tr_\tau \sim p_\theta(\tr_\tau\mid \ts_\tau)$ and the next action $\ta_\tau \sim q_\eta (\ta_\tau\mid \ts_\tau)$. We repeat this process until we have an imagined trajectory $\{\ts_\tau, \ta_\tau, \tr_\tau\}_{\tau=t}^{t+H}$.

\subsection{Value Estimation of Dreamer}
Dreamer estimates the value $\tilde{V}(s_\tau)$ of imagined trajectories using the following equations:
\begin{align*}
    \tV_N^k(\ts_\tau) &= \mathbb{E}_{p_\theta, q_\eta}\left( \sum\limits_{n=\tau}^{h-1} \gamma^{n-\tau} \tr_n + \gamma^{h-\tau} v_\psi (\ts_h)\right), \quad\mbox{where } h=\min (\tau + k, t+H)\\
    \tV_\lambda(\ts_\tau) &= (1- \lambda) \sum\limits_{n=1}^{H-1}\lambda^{n-1}\tV_N^n(\ts_\tau) + \lambda^{H-1}\tV_N^H(\ts_\tau)
\end{align*}
$\tV_N^k(\ts_\tau)$ estimates the value of $\ts_\tau$ using the rewards of $k$ steps of rollouts and the value function estimate $v_\psi$ of the last state. Dreamer ues $\tV_\lambda(\ts_\tau)$ as the final value estimation, which is an exponentially-weighted average of different $k$-step rollouts to tradeoff the bias and variance.

\subsection{Latent Guided MPC}

Originally, for each state $s_t$, the actor network $q_\eta(a_t\mid s_t)$ generates the action which maximizes the long-horizon accumulated reward. However, the approximation highly depends on the quality of the learned world model and might have difficulties approximating complex policies. Most importantly, it lacks the reasoning ability to adapt to variable environments. 

We use the shooting method for MPC with differentiable world model. Specifically, we use stochastic gradient ascent to optimize the action sequences to output high accumulated reward.
During execution, for each obsevation $o_t$, previous state $s_{t-1}$ and action $a_{t-1}$, we encode / propose the current state by $q_\phi(s_t\mid o_{\leq t}, a_{\leq t})$. Next, we perform latent imagination and sample the imagined trajectories $\{\ts_\tau, \ta_\tau, \tr_\tau\}_{\tau=t}^{t+H}$ and estimate $\tV_\lambda (s_t)$. As $\tV_\lambda (s_t)$ is computed using predicted rewards and value estimations, which are conditioned on the action sequences, we can backpropagate the gradients from $\tV_\lambda (s_t)$ to the actions. We update the actions by
\begin{equation}
    \ta_\tau ' = \ta_\tau + \nabla_{\ta_\tau} \tV_\lambda (s_t)\nonumber
\end{equation}
We repeat this for all actions and return the first action after update.

Our latent guided MPC is similar to the planning algorithm used in DPI-Net~\citep{li2018learning}. The difference is that DPI-Net requires a pre-defined observation of the goal to compute the loss, whereas CVRL directly maximizes the accumulated reward and alleviate this assumption.

\section{Implementation Details}
\subsection{Negative Sample Selection}\label{sect:neg_sel}
We adopt a simple strategy to generate negative samples. We sample a batch of sequences $\{o_{1:T}^{(i)}, a_{1:T}^{(i)}, r_{1:T}^{(i)}\}_{i=1}^B$ from a replay buffer, where $T$ is the sequence length and $B$ is the batch size. For each state-observation pair $(s_t, o_t)$, we treat the other $B*T - 1$ observations $\{o'\}$ in the same batch as negative samples. An intuition of this choice is that: 1) by contrasting $(s_t^{(i)}, o_t^{(i)})$ with $(s_t^{(i)}, o_{t'}^{(j)})$ where $j\neq i$ and $t'\in [1, T]$, CELBO learns to identify invariant features of the task given variable visual features; 2) by contrasting $(s_t^{(i)}, o_t^{(i)})$ with $(s_t^{(i)}, o_{t'}^{(i)})$ where $t'\neq t$, CELBO learns to model the temporal dynamics of the task. We found this simple strategy works well in practice. 

\subsection{Hardware and Software.} We train all models on single NVidia RTX 2080Ti GPUs with Intel Xeon Gold 5220 CPU @ 2.20GHz. We implement all models with Tensorflow 2.2.0 and Tensorflow Probability 0.10.0. Specifically, our code is developed based on the official Tensorflow implementation of Dreamer, but heavily modified. We use the official implementation of Dreamer as our baseline, and we use the SAC implementation of OpenAI baselines. For all methods, we share certain structure, including the encoder, RSSM model and the actor-critic networks to make it a fair comparison.

\subsection{Observation Encoder.} We use an encoder of 4 convolutional layers for image observations, which have a fixed kernel size of 4 with increasing channel numbers: 32, 65, 128, 256. We do not encode the actions again and directly concatenate it with the states.

\subsection{RSSM.} We use a stochastic state $s_t$ with size 30 and a deterministic state with size 200. The deterministic update function is parameterized using a GRU and the for the stochastic part, we learn the mean and variance of $s_t$ using two fully connected layers with size 200 and 30.

\subsection{Contrastive Learning.} In contrastive learning, we learn the compatibility between state $s_t$ and observation $o_t$ with a function $f_\theta (s_t, o_t)$. In our implementation, we first encode both $o_t$ and $s_t$ by two separate fully connected layers with size $200$, then we compute the value of $f_\theta (s_t, o_t) = \exp(z_t^\mathrm{T}W_\theta s_t')$, where $z_t$ and $s_t'$ are the embeddings of the observation and the state, and $w_\theta$ is a $200\times 200$ matrix.

\subsection{Actor-Critic.} For the actor network, we use 4 fully connected layer which takes in the concatenation of $s_t$ and $h_t$ as input, with intermediate hidden dimension of 400, and output the corresponding action, with tanh as the activation function. Specifically, a transformed distribution is used to achieve differentiable sampling. For the value network, 3 fully connected layers are used with hidden dimension of 400 and output dimension of 1. In addition, SAC needs additional Q-value network during training. For models needs SAC, we use 2 Q-value network with similar structure, except that the input is a concatenation of $s_t$, $h_t$ and $a_t$.

\subsection{Model Learning.} We train CVRL by 4 separate optimizers for different part of the network: model optimizer, value optimizer, actor optimizer and SAC optimizer. For all optimizers, we use Adam optimizer in our implementation with different learning rate. Model optimizer updates all contrastive variational world model dynamics by representation learning defined in Eqn.~\ref{eqn:celbo} with learning rate $6\times 10^{-4}$; value optimizer updates only value network parameters with learning rate $8\times 10^{-5}$; actor optimizer updates the actor parameters with learning rate $8 \times 10^{-5}$; SAC optimizer updates the actor parameters and the two Q-value network parameters with learning rate $8\times 10^{-5}$.

\subsection{Latent Guided MPC.} In latent guided MPC, we unroll for 15 steps and update the actions by standard SGD with learning rate 0.003.

\clearpage
\section{Additional Visualizations}
\subsection{Natural Mujoco Tasks}
\begin{figure}[!htb]
	\centering
	\begin{tabular}{c c c}
		\includegraphics[width=0.31\linewidth]{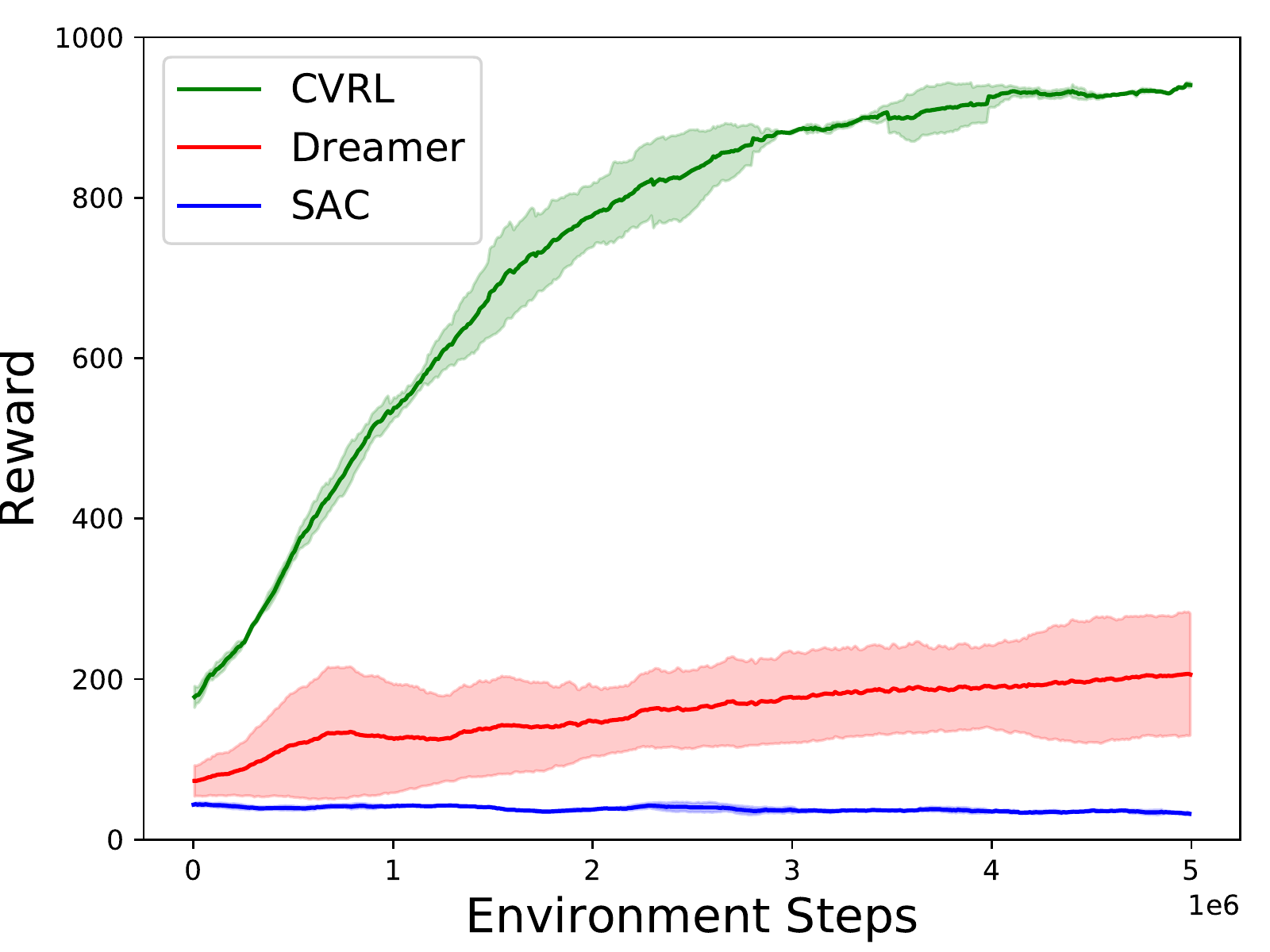} &
		\includegraphics[width=0.31\linewidth]{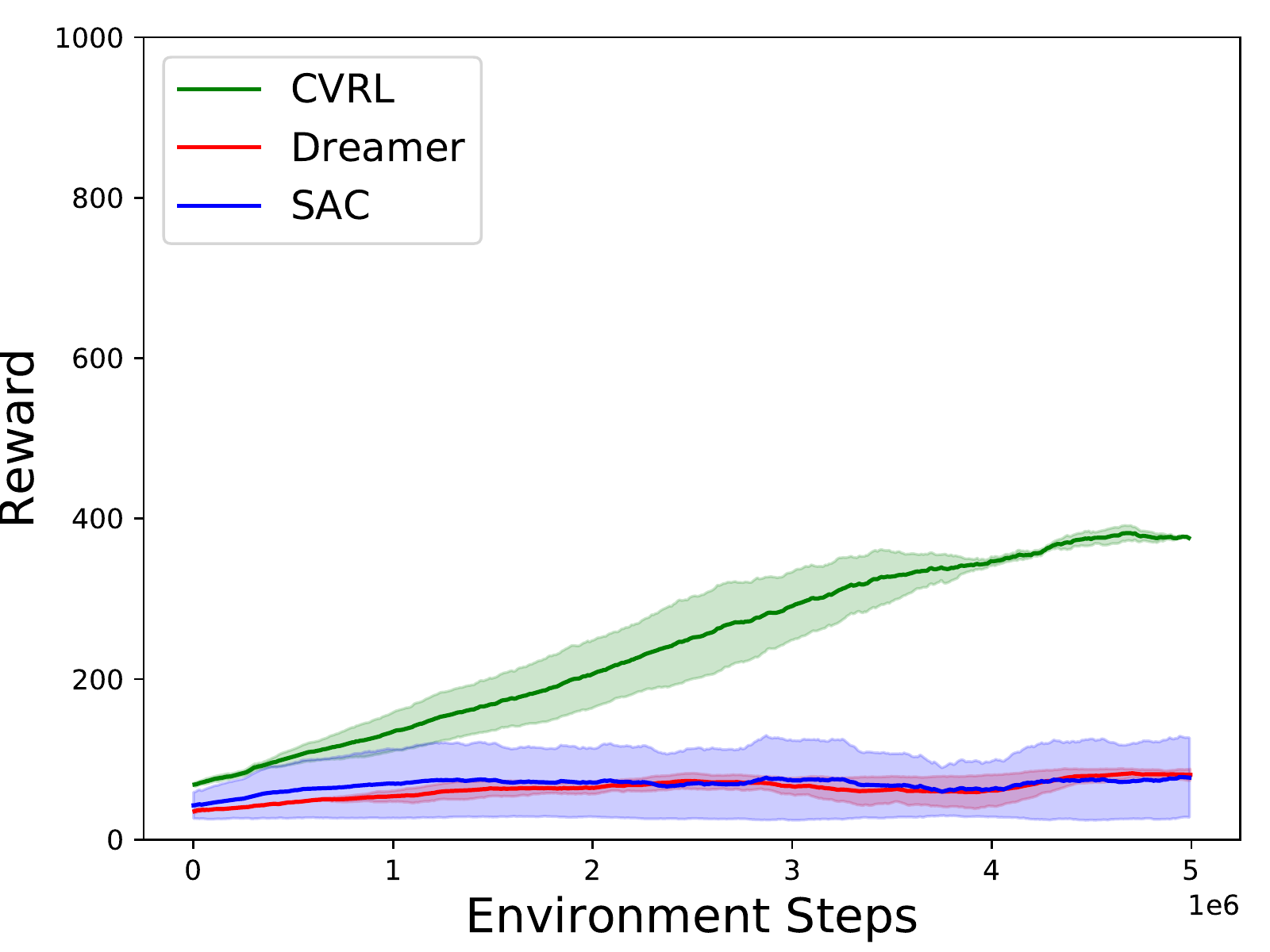}&
		\includegraphics[width=0.31\linewidth]{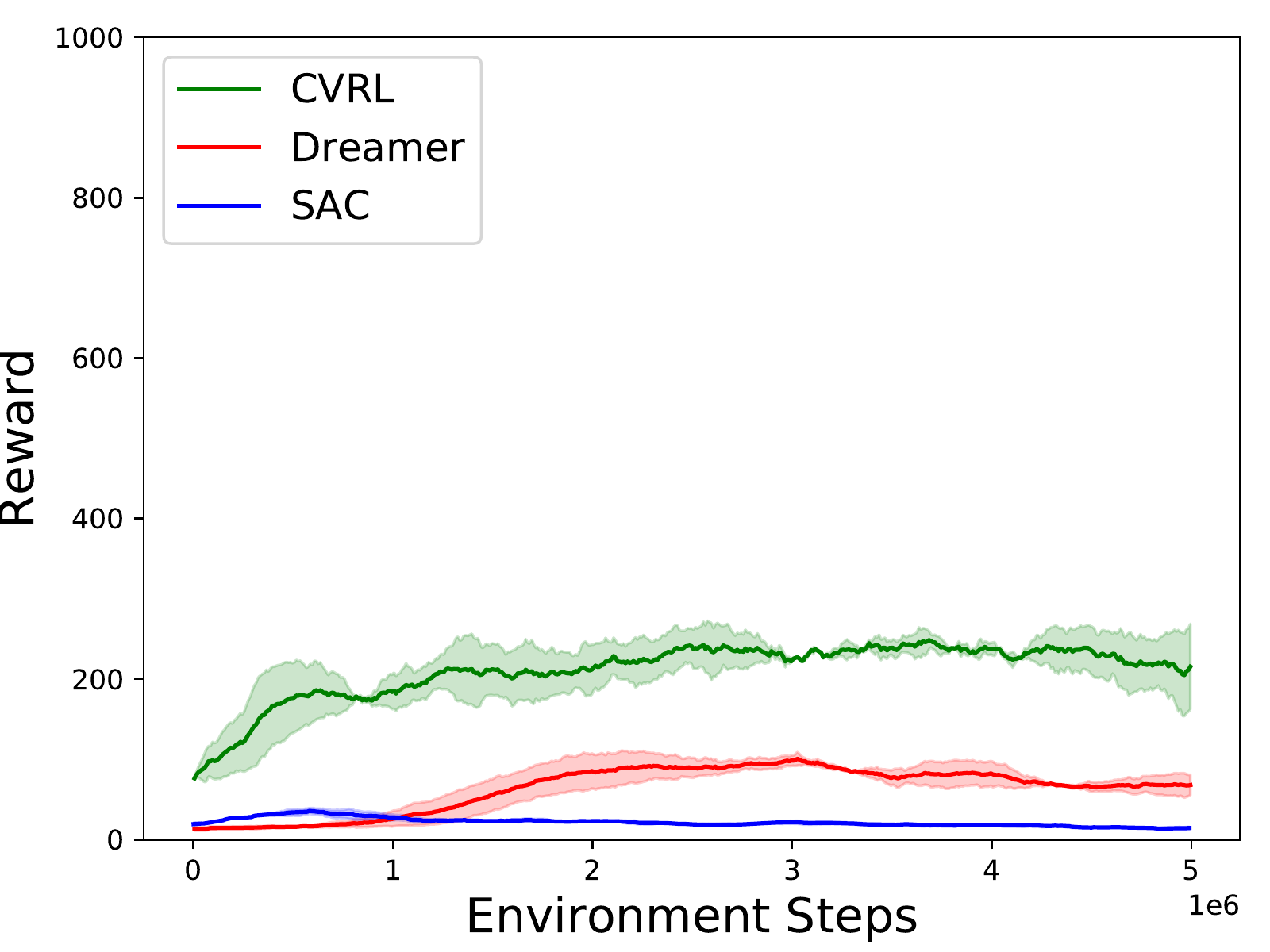}\\
		(a) Natural Walker Walk & (b) Natural Walker Run & (c) Natural Cheetah Run\\
		\includegraphics[width=0.31\linewidth]{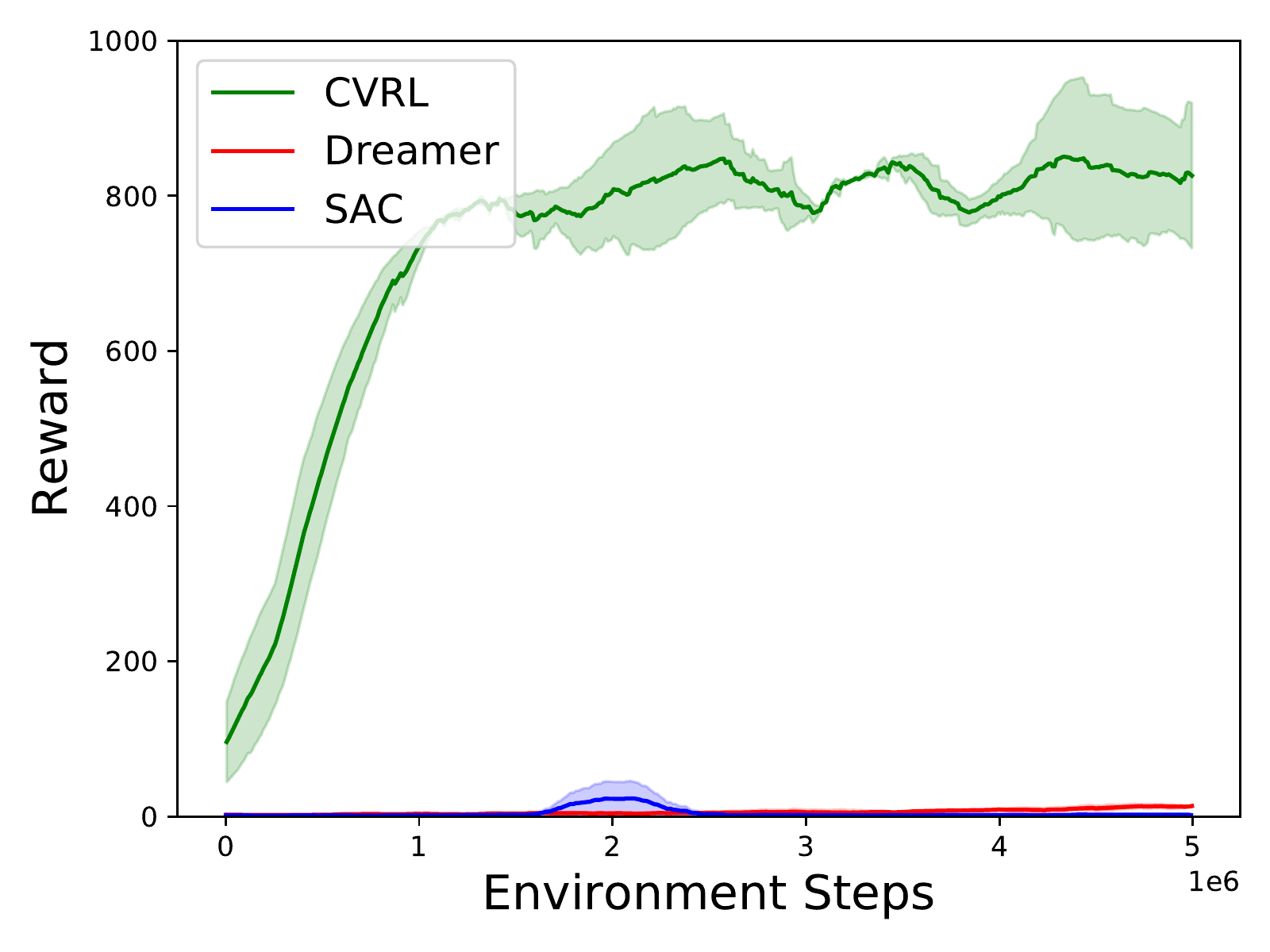} &
		\includegraphics[width=0.31\linewidth]{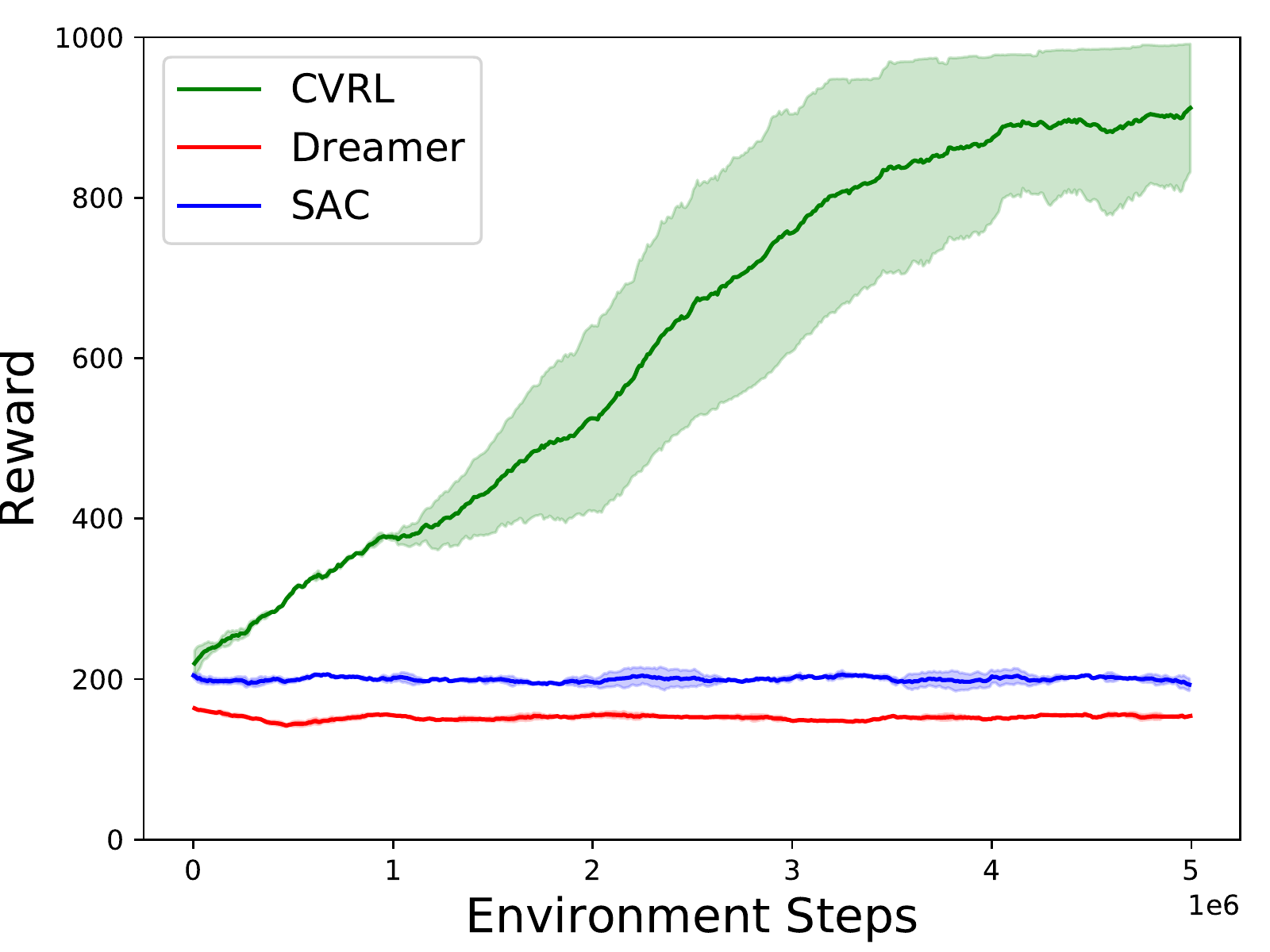}&
		\includegraphics[width=0.31\linewidth]{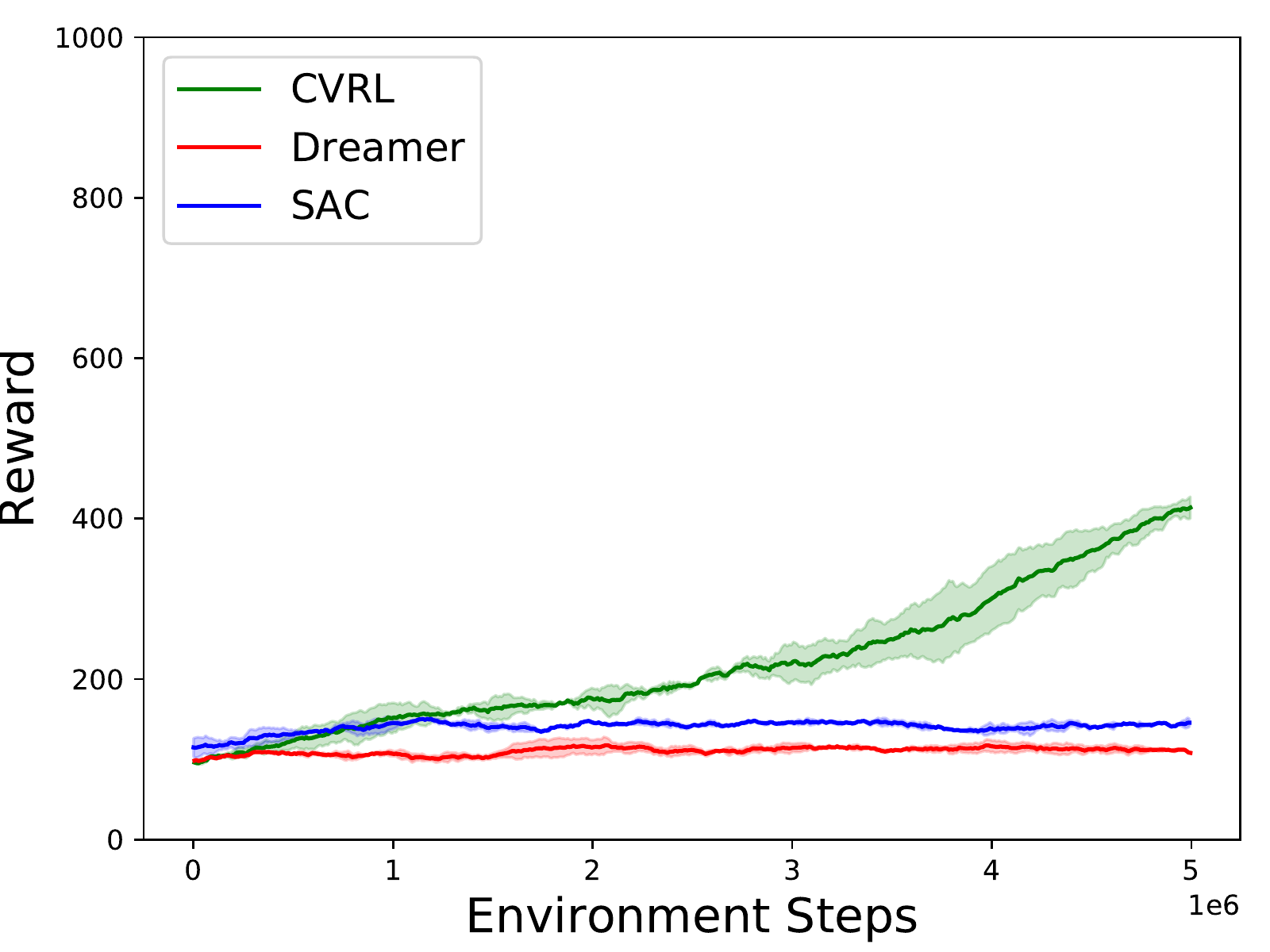}\\
		(d) Natural Finger Spin& (e) Natural Cartpole Balance& (f) Natural Cartpole Swingup\\
		\includegraphics[width=0.31\linewidth]{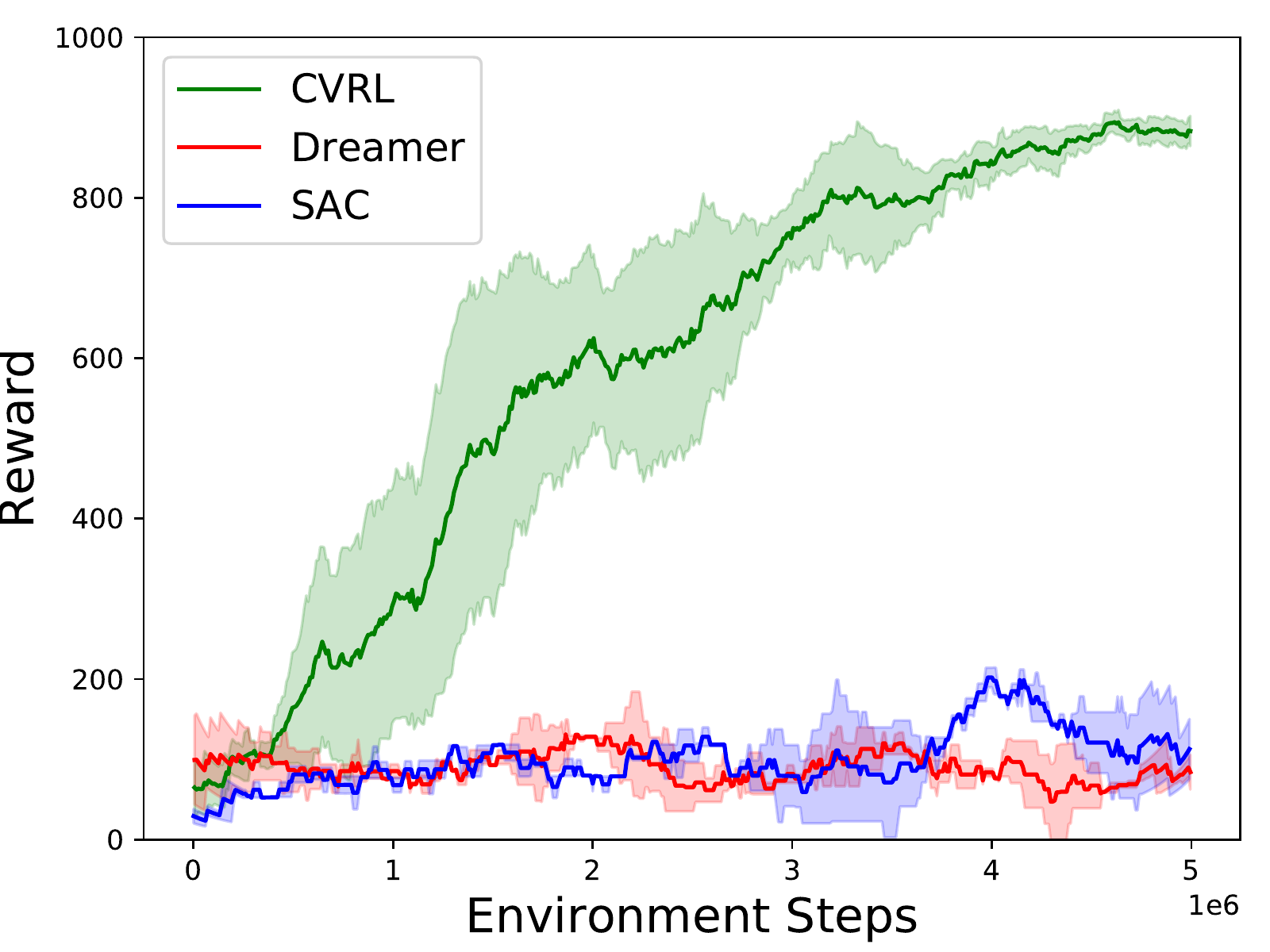} &
		\includegraphics[width=0.31\linewidth]{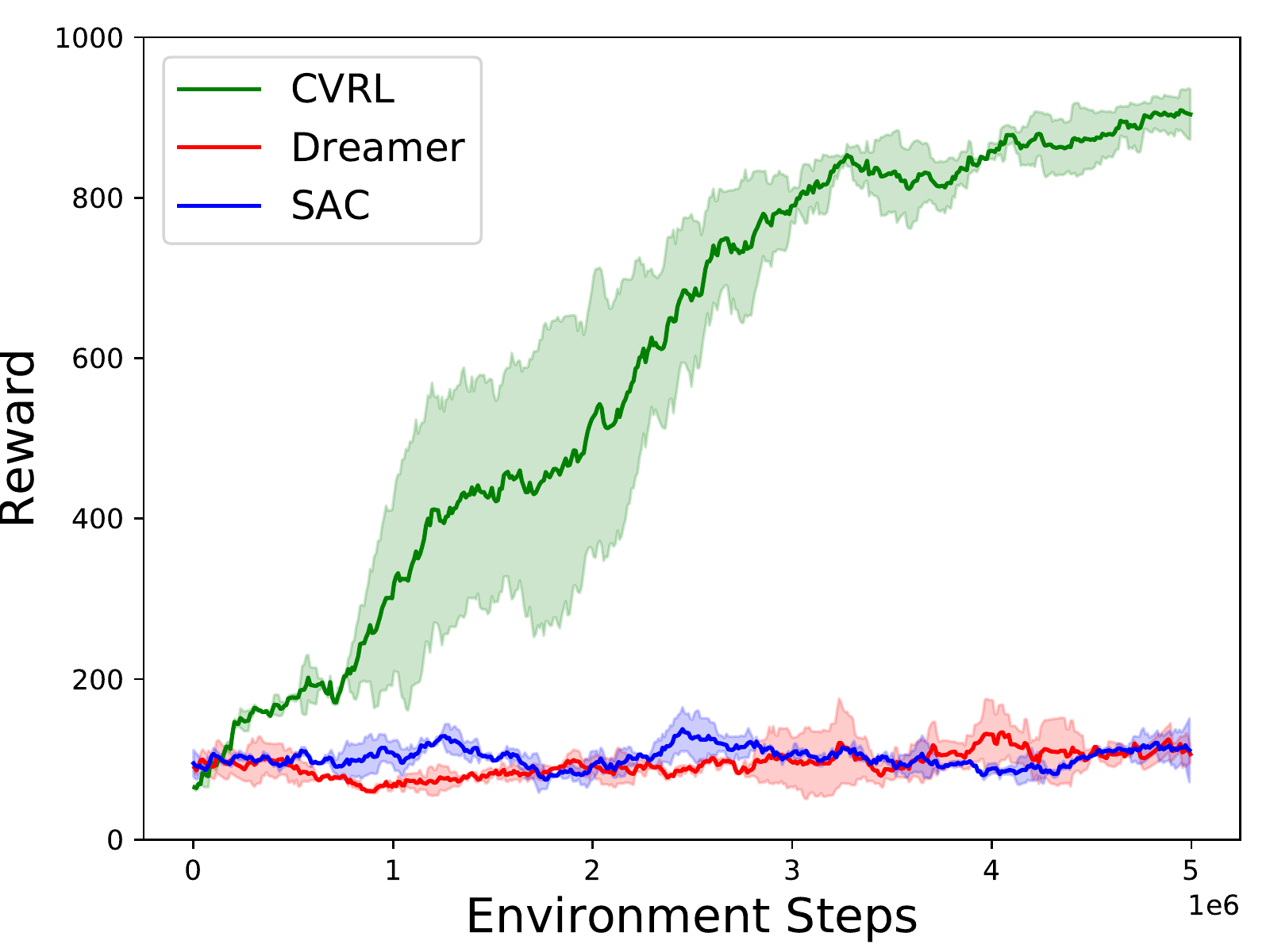}&
		\includegraphics[width=0.31\linewidth]{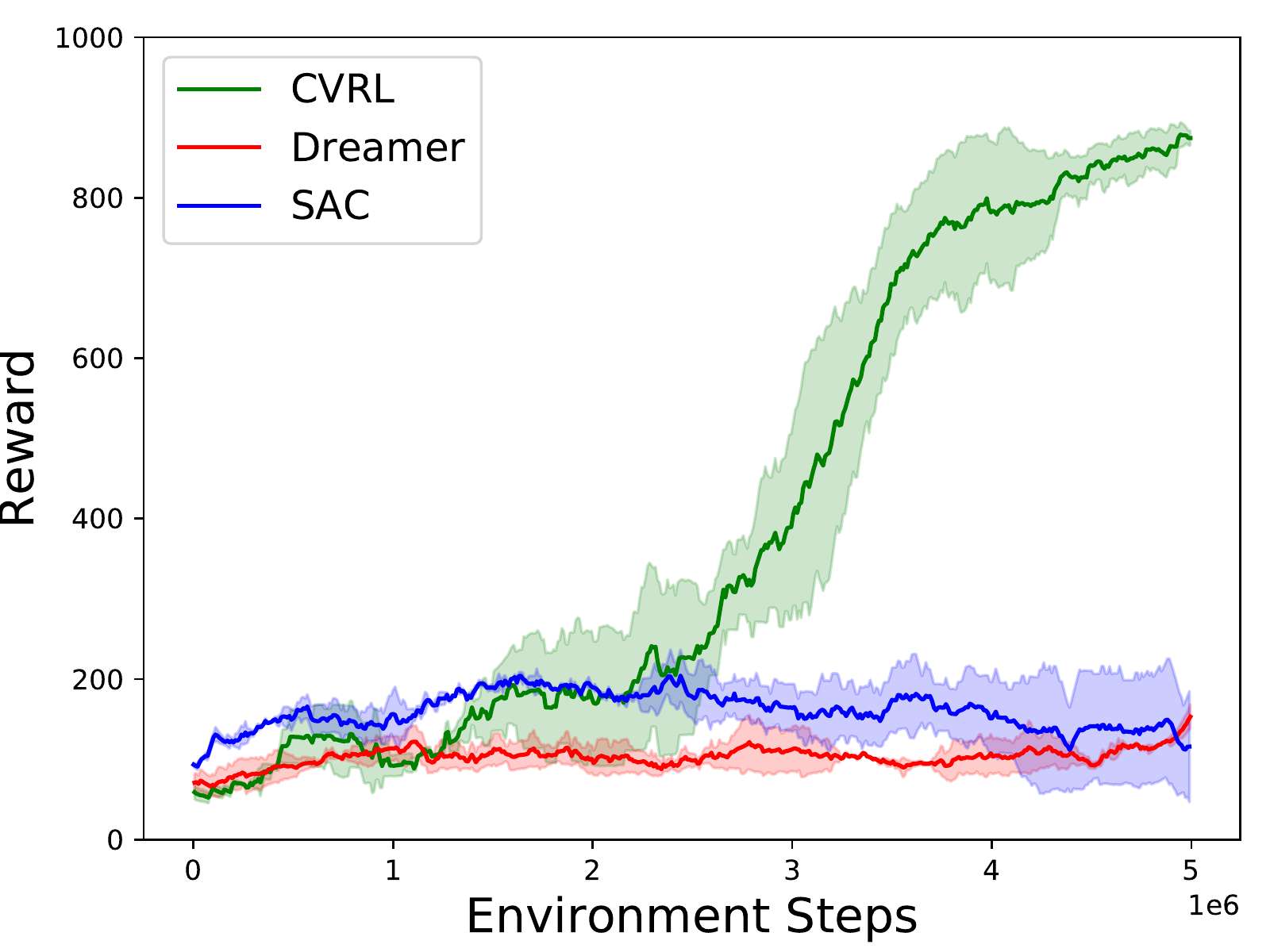}\\
		(g) Natural Cup Catch& (h) Natural Reacher Easy& (i) Natural Quadruped Walk\\
		\includegraphics[width=0.31\linewidth]{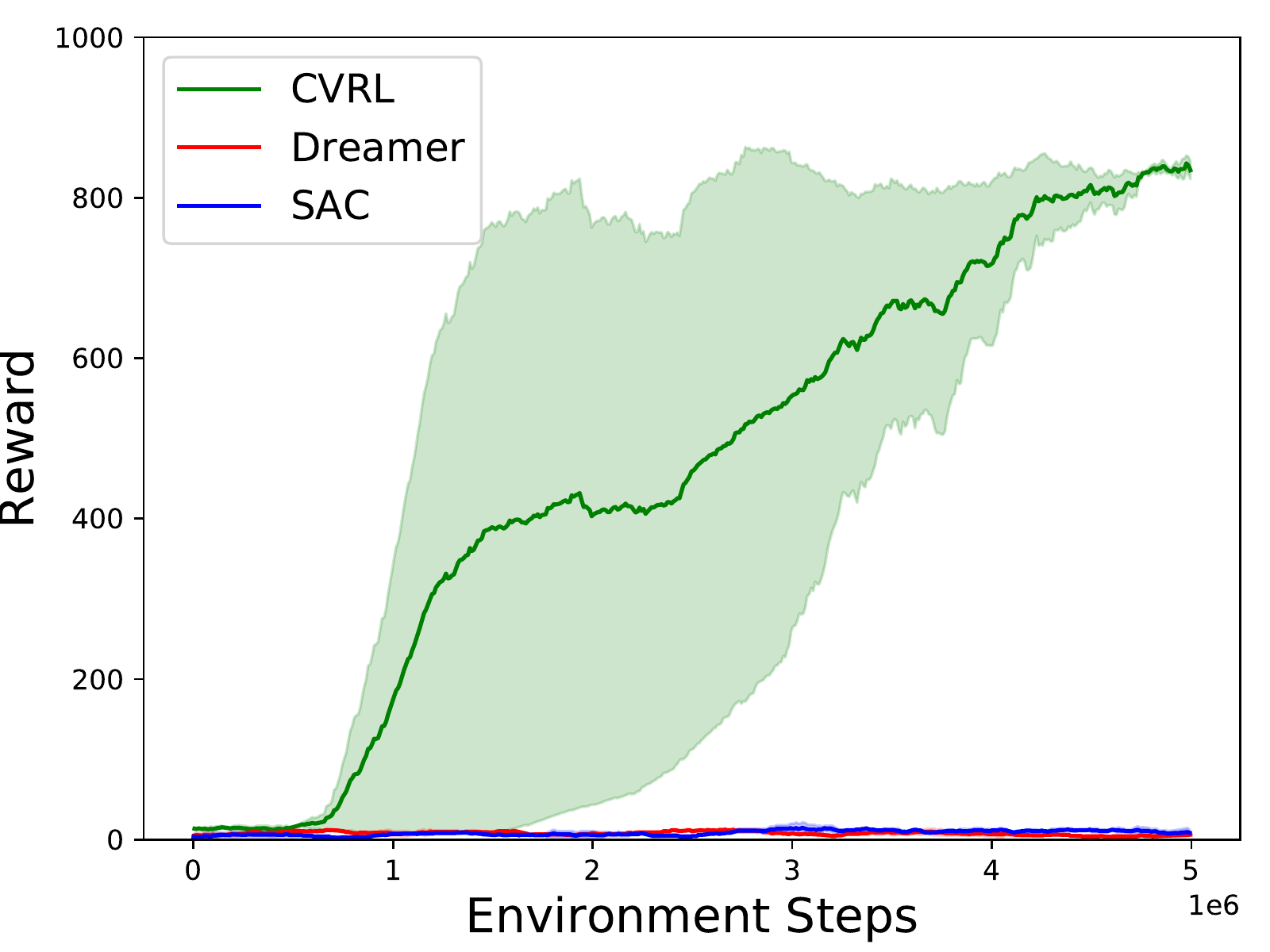} & &\\
		(j) Natural Pendulum Swingup& &\\
		
	\end{tabular}
\end{figure}

\clearpage
\subsection{Standard Mujoco Tasks}
\begin{figure}[!htb]
	\centering
	\begin{tabular}{c c c}
		\includegraphics[width=0.31\linewidth]{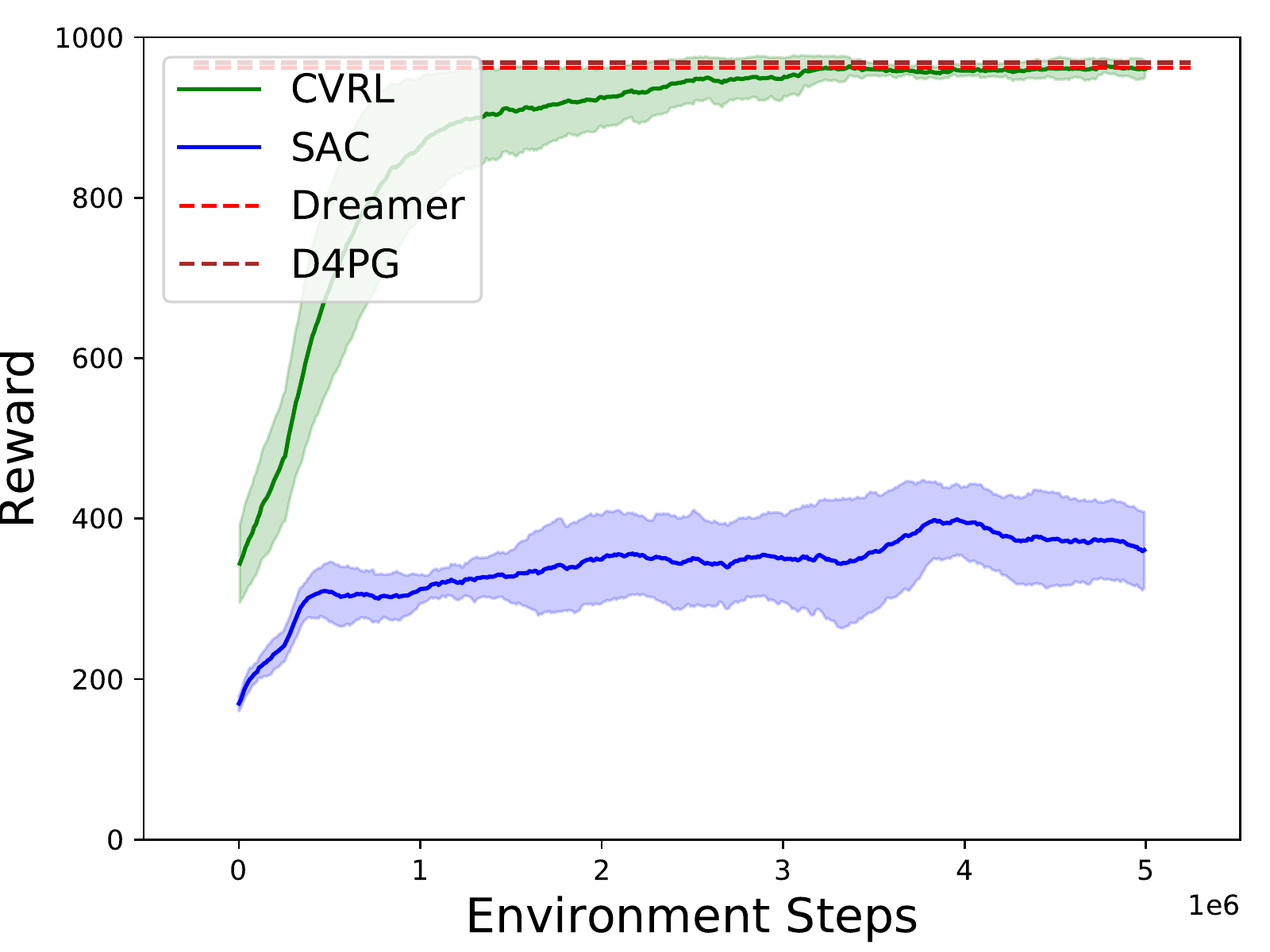} &
		\includegraphics[width=0.31\linewidth]{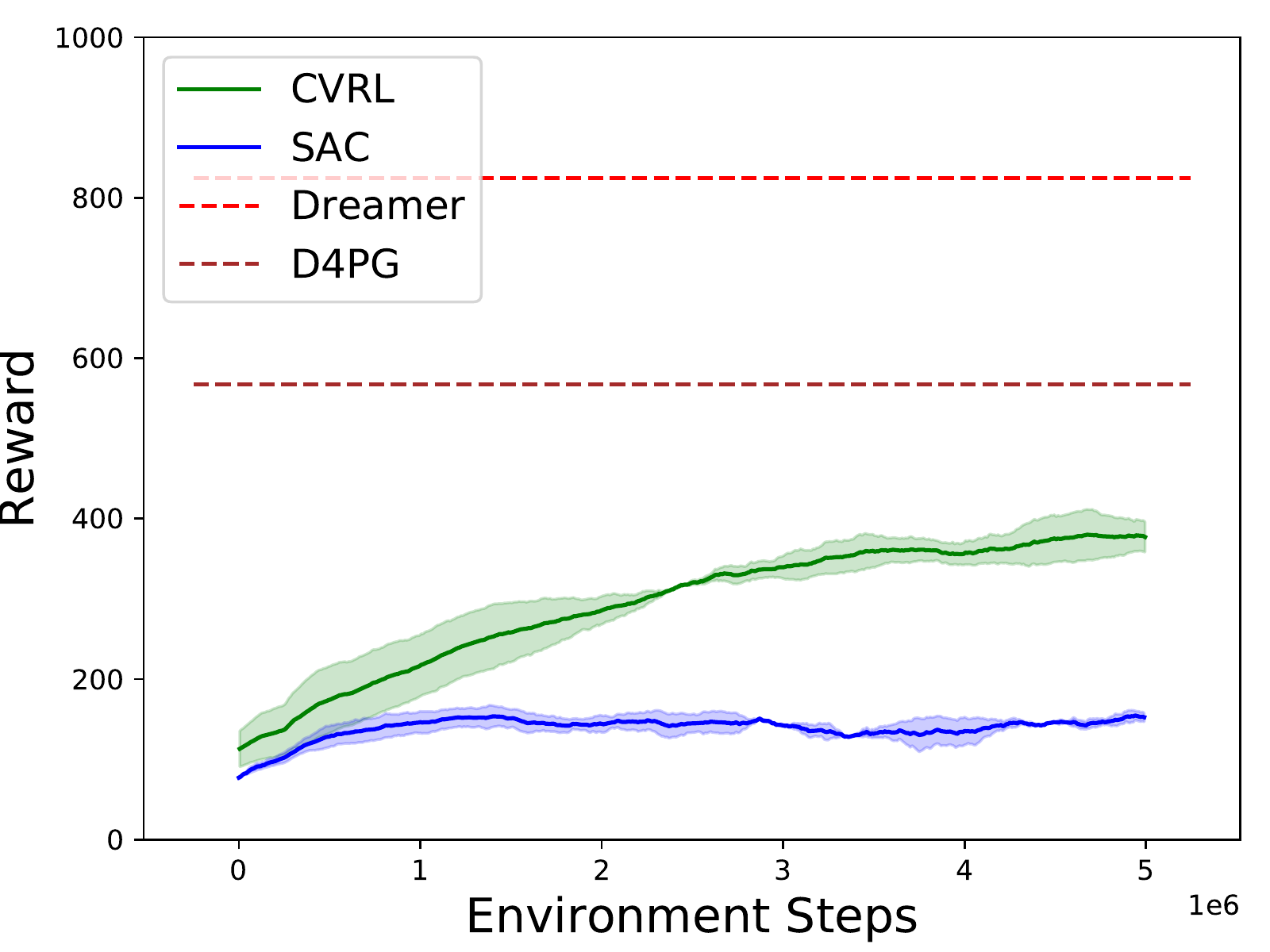}&
		\includegraphics[width=0.31\linewidth]{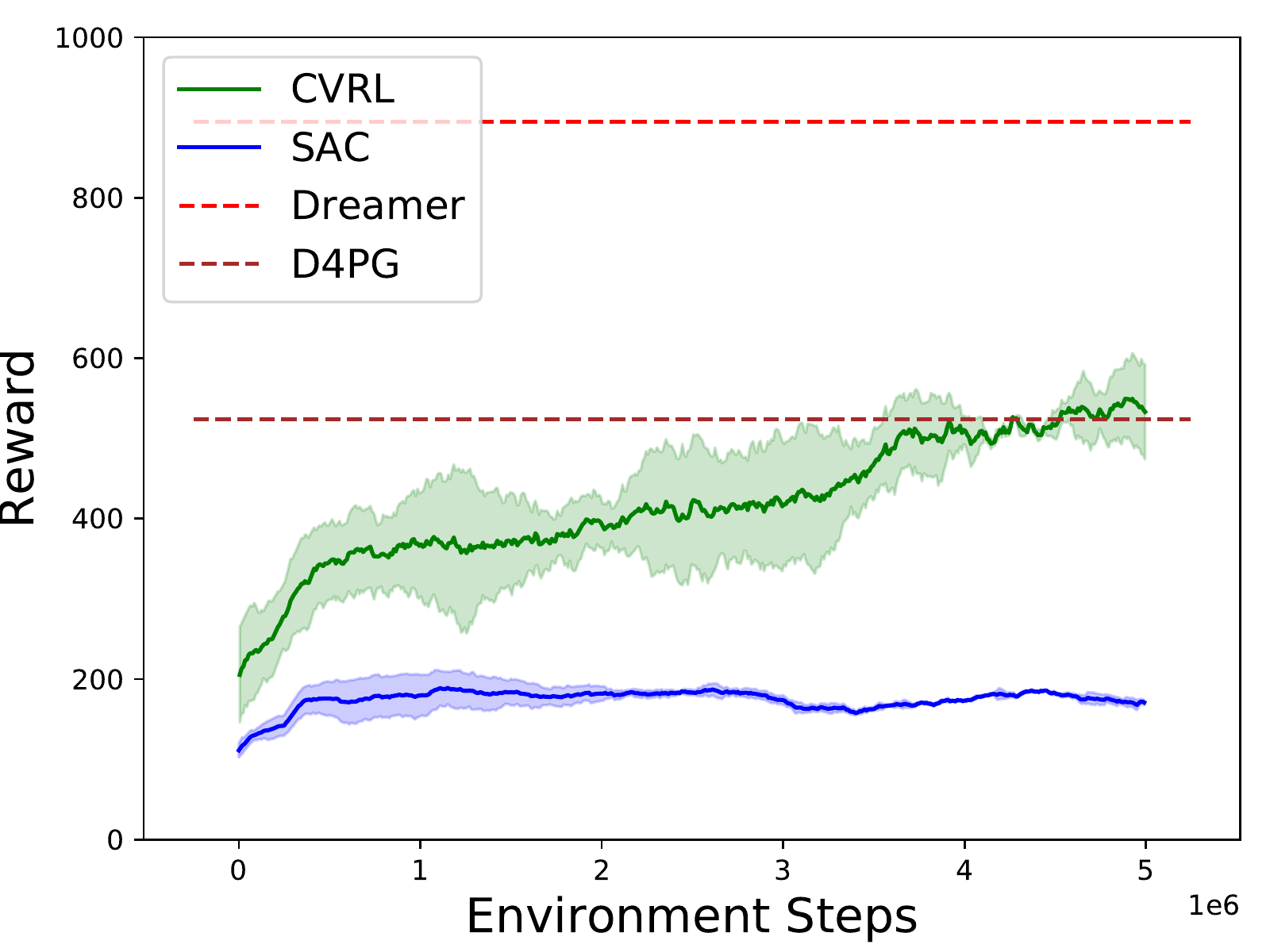}\\
		(a) Standard Walker Walk & (b) Standard Walker Run & (c) Standard Cheetah Run\\
		\includegraphics[width=0.31\linewidth]{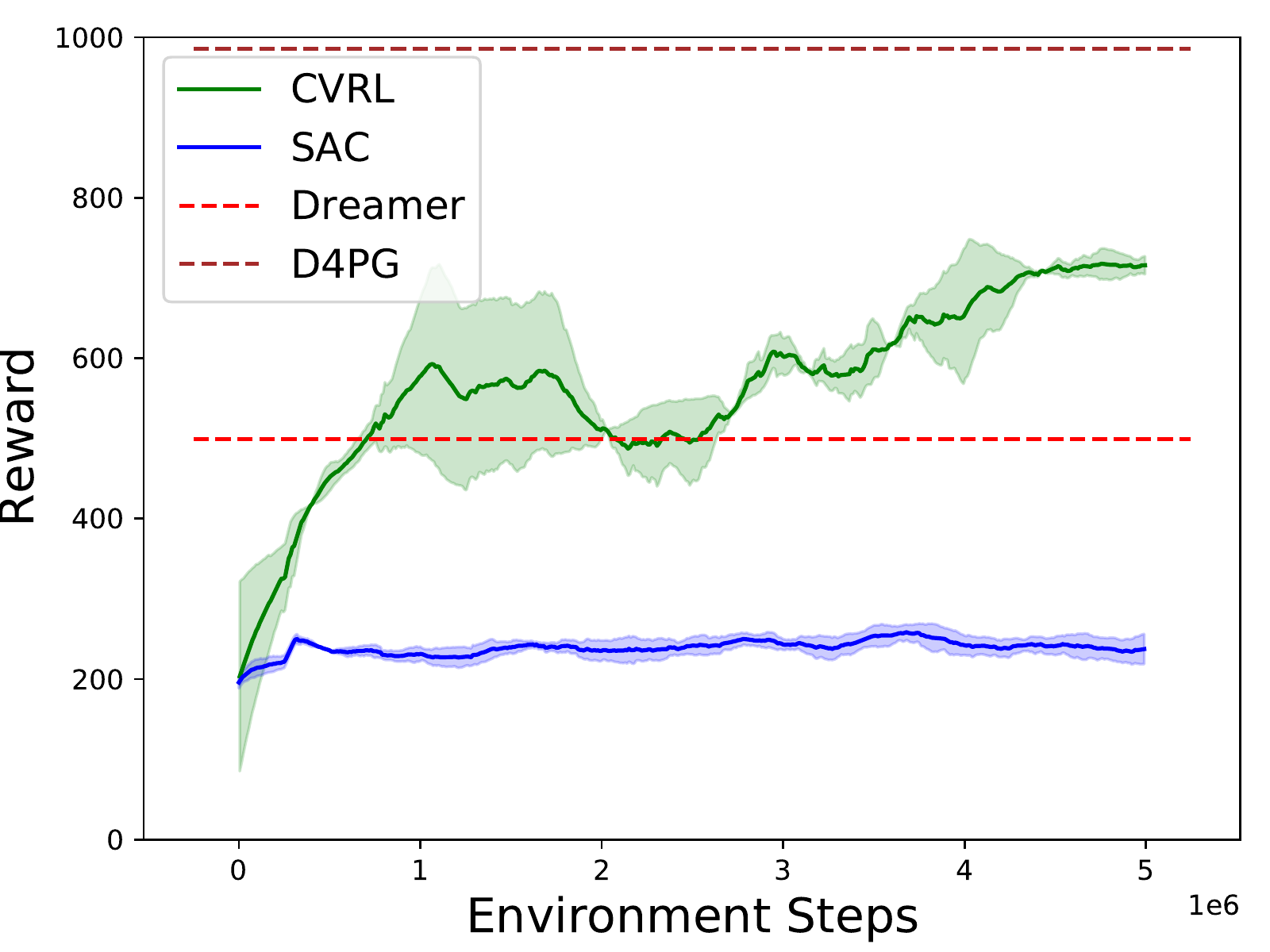} &
		\includegraphics[width=0.31\linewidth]{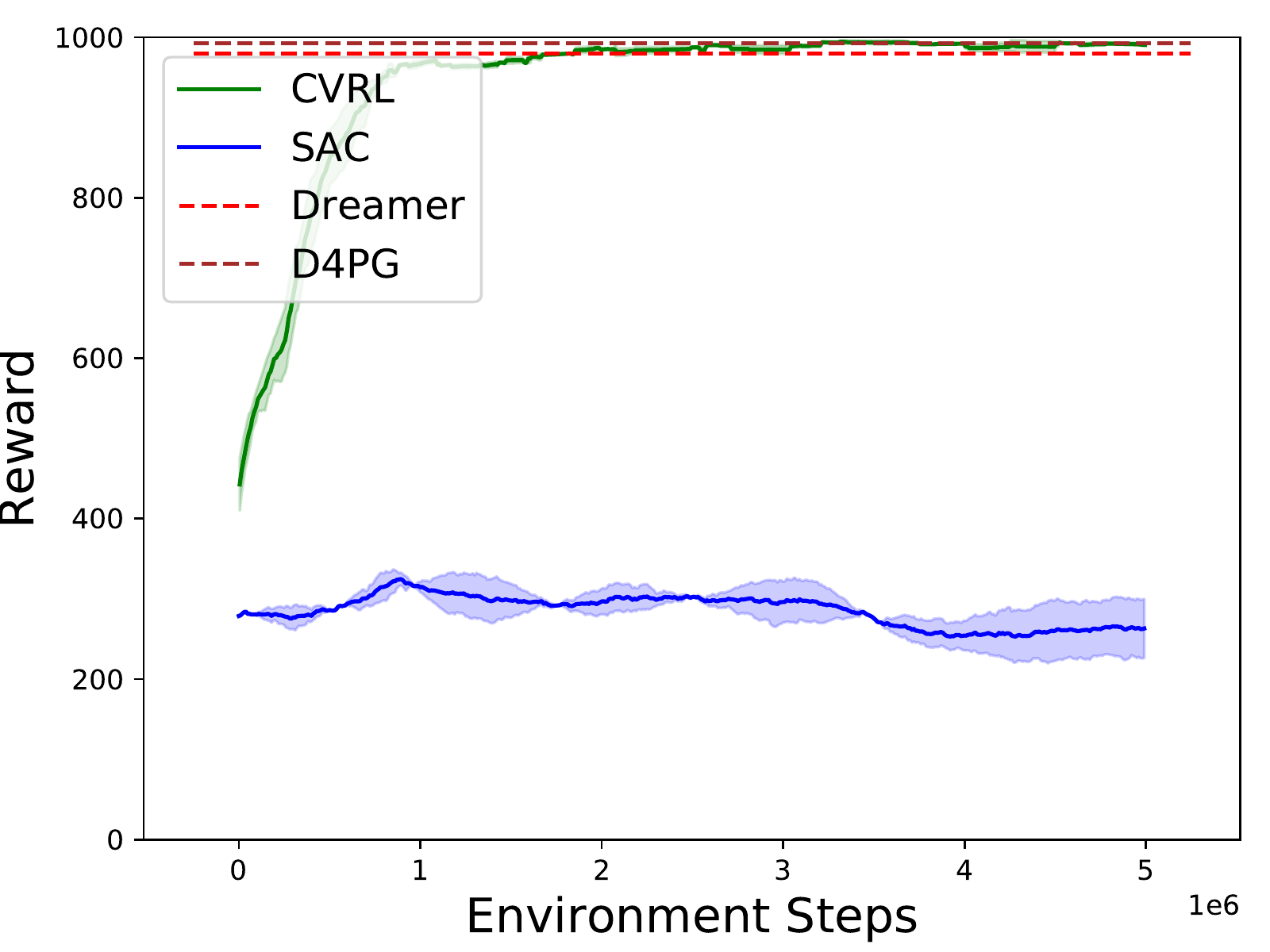}&
		\includegraphics[width=0.31\linewidth]{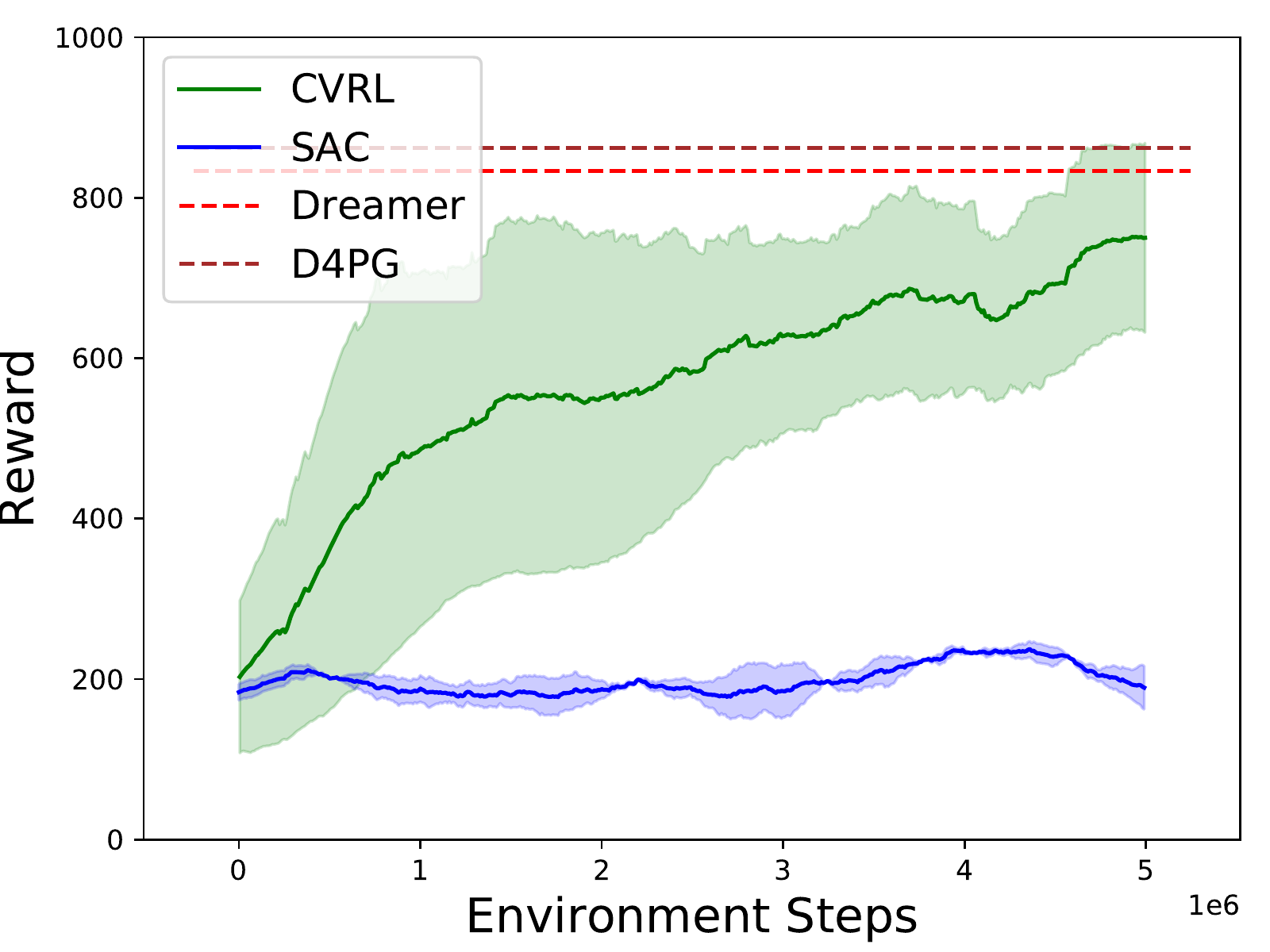}\\
		(d) Standard Finger Spin& (e) Standard Cartpole Balance& (f) Standard Cartpole Swingup\\
		\includegraphics[width=0.31\linewidth]{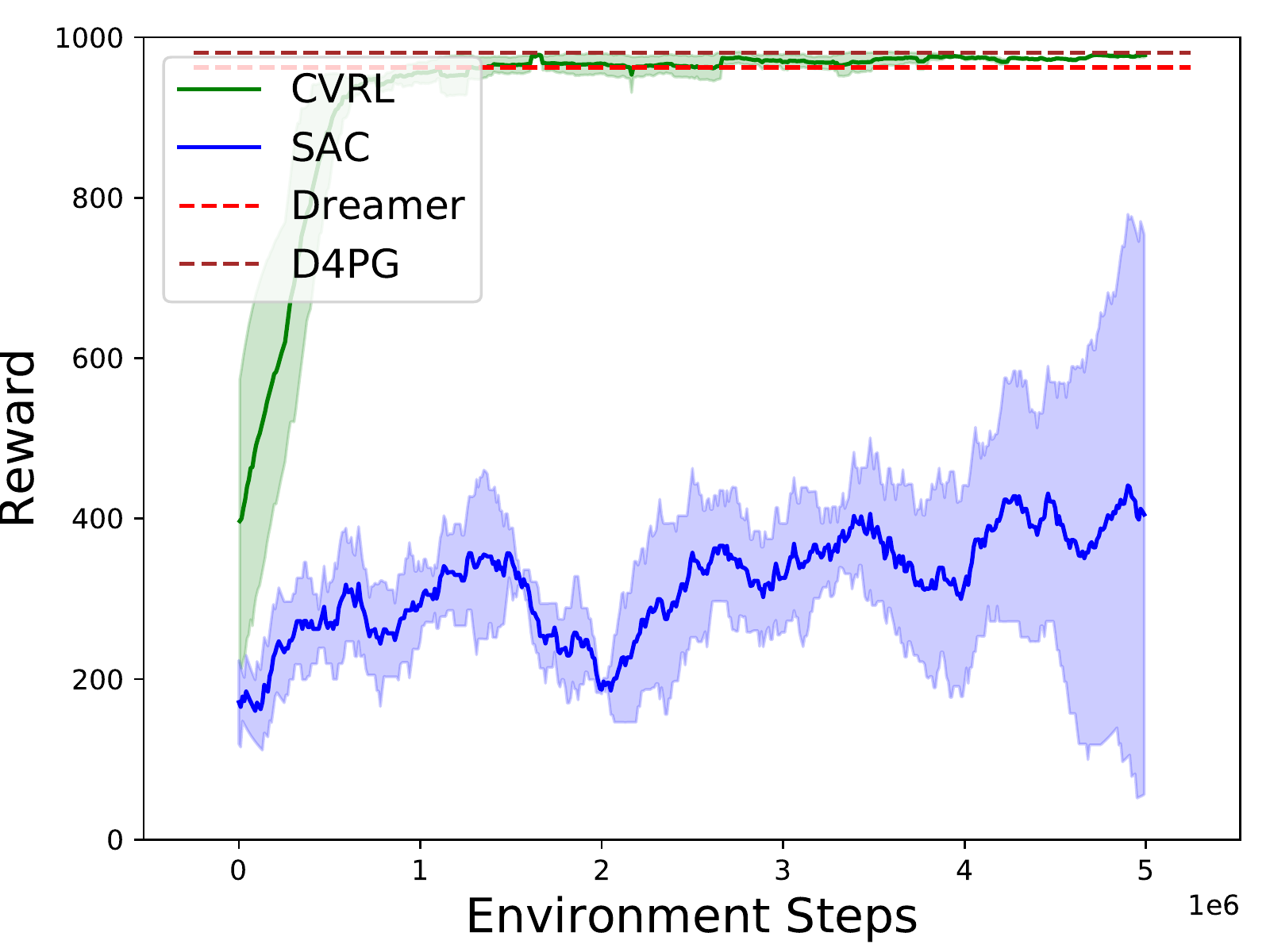} &
		\includegraphics[width=0.31\linewidth]{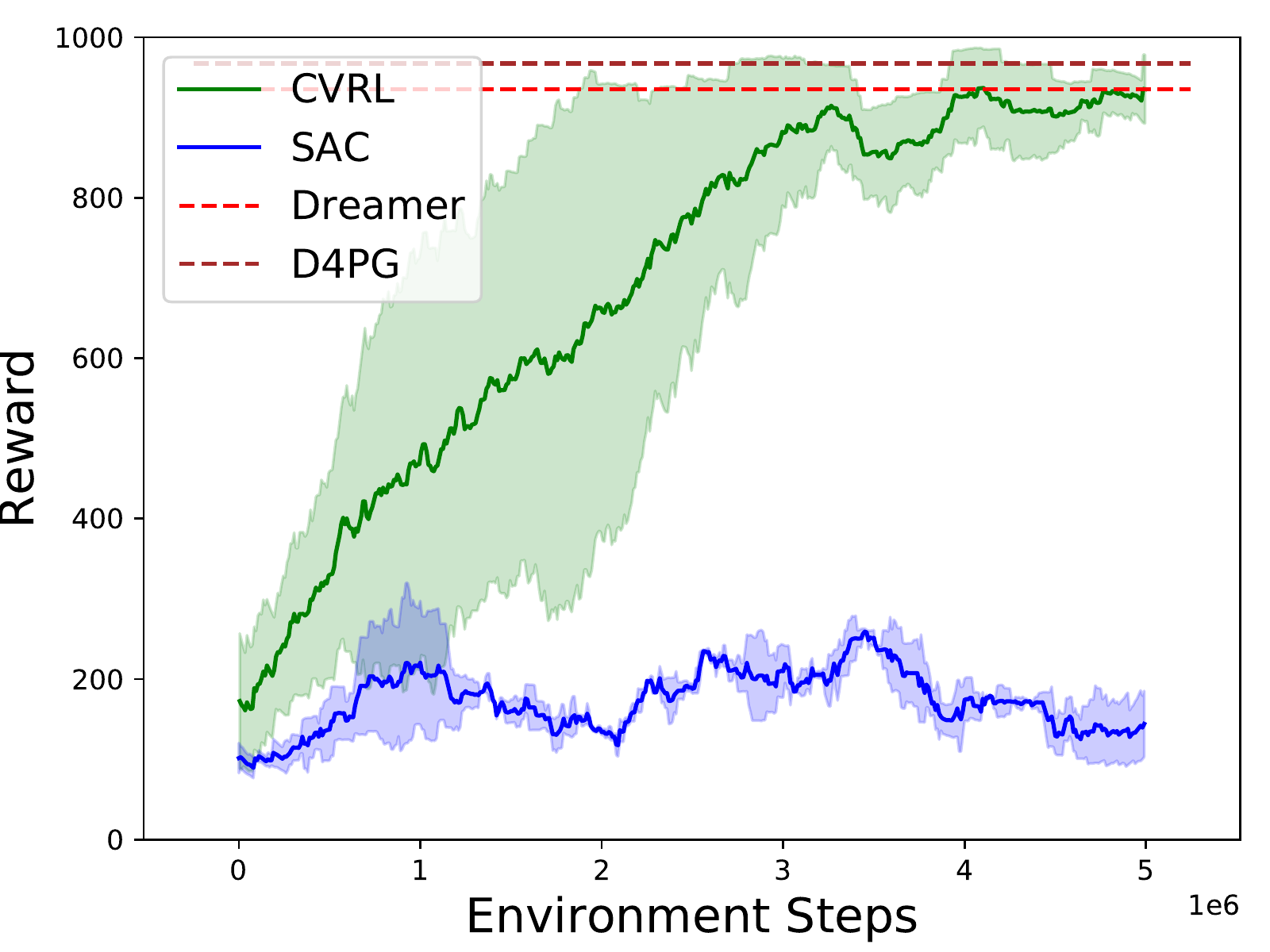}&
		\includegraphics[width=0.31\linewidth]{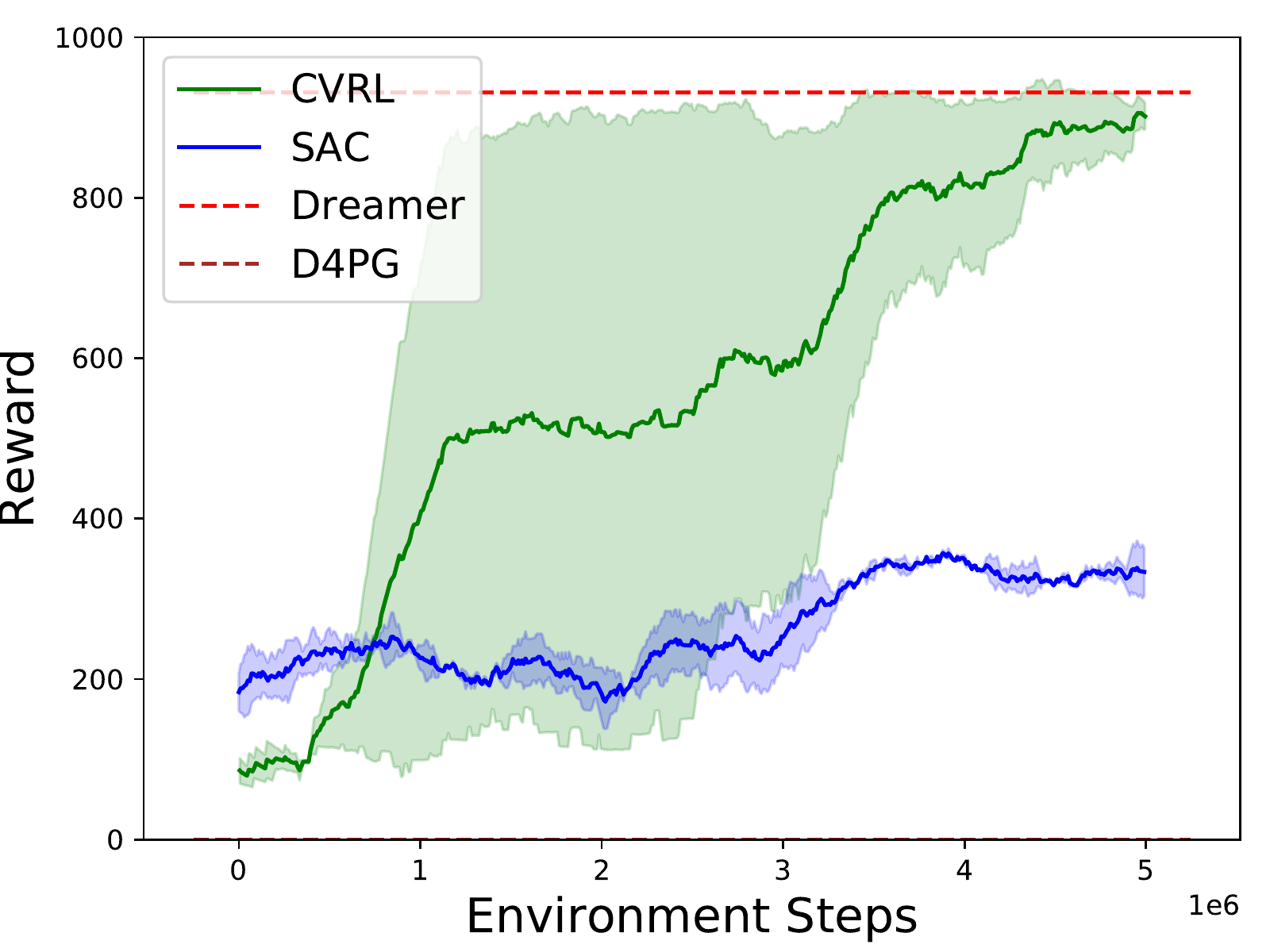}\\
		(g) Standard Cup Catch& (h) Standard Reacher Easy& (i) Standard Quadruped Walk\\
		\includegraphics[width=0.31\linewidth]{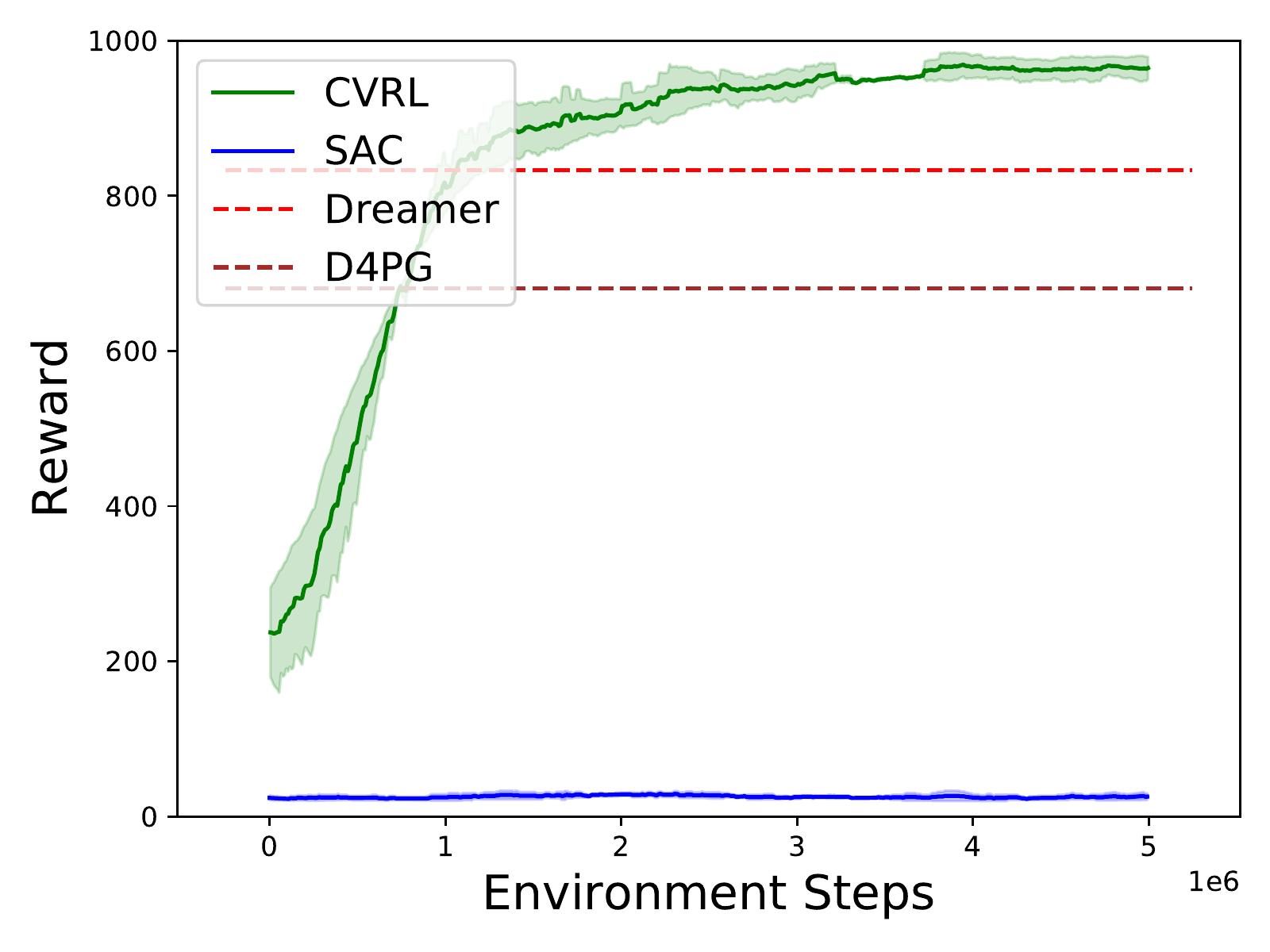} & &\\
		(j) Standard Pendulum Swingup& &\\
		
	\end{tabular}
\end{figure}


\begin{thebibliography}{39}
\providecommand{\natexlab}[1]{#1}
\providecommand{\url}[1]{\texttt{#1}}
\expandafter\ifx\csname urlstyle\endcsname\relax
  \providecommand{\doi}[1]{doi: #1}\else
  \providecommand{\doi}{doi: \begingroup \urlstyle{rm}\Url}\fi

\bibitem[Mnih et~al.(2013)Mnih, Kavukcuoglu, Silver, Graves, Antonoglou,
  Wierstra, and Riedmiller]{mnih2013playing}
V.~Mnih, K.~Kavukcuoglu, D.~Silver, A.~Graves, I.~Antonoglou, D.~Wierstra, and
  M.~Riedmiller.
\newblock Playing atari with deep reinforcement learning.
\newblock \emph{arXiv preprint arXiv:1312.5602}, 2013.

\bibitem[Silver et~al.(2017)Silver, Schrittwieser, Simonyan, Antonoglou, Huang,
  Guez, Hubert, Baker, Lai, Bolton, et~al.]{silver2017mastering}
D.~Silver, J.~Schrittwieser, K.~Simonyan, I.~Antonoglou, A.~Huang, A.~Guez,
  T.~Hubert, L.~Baker, M.~Lai, A.~Bolton, et~al.
\newblock Mastering the game of go without human knowledge.
\newblock \emph{Nature}, 2017.

\bibitem[Levine et~al.(2016)Levine, Finn, Darrell, and Abbeel]{levine2016end}
S.~Levine, C.~Finn, T.~Darrell, and P.~Abbeel.
\newblock End-to-end training of deep visuomotor policies.
\newblock \emph{The Journal of Machine Learning Research}, 17\penalty0
  (1):\penalty0 1334--1373, 2016.

\bibitem[Kahn et~al.(2018)Kahn, Villaflor, Ding, Abbeel, and
  Levine]{kahn2018self}
G.~Kahn, A.~Villaflor, B.~Ding, P.~Abbeel, and S.~Levine.
\newblock Self-supervised deep reinforcement learning with generalized
  computation graphs for robot navigation.
\newblock In \emph{2018 IEEE International Conference on Robotics and
  Automation (ICRA)}, pages 1--8. IEEE, 2018.

\bibitem[Bosworth et~al.(2015)Bosworth, Kim, and Hogan]{bosworth2015super}
W.~Bosworth, S.~Kim, and N.~Hogan.
\newblock The mit super mini cheetah: A small, low-cost quadrupedal robot for
  dynamic locomotion.
\newblock In \emph{2015 IEEE International Symposium on Safety, Security, and
  Rescue Robotics (SSRR)}, pages 1--8. IEEE, 2015.

\bibitem[Allen and Koomen(1983)]{allen1983planning}
J.~F. Allen and J.~A. Koomen.
\newblock Planning using a temporal world model.
\newblock In \emph{Proceedings of the Eighth international joint conference on
  Artificial intelligence-Volume 2}, pages 741--747, 1983.

\bibitem[Basye et~al.(1992)Basye, Dean, Kirman, and Lejter]{basye1992decision}
K.~Basye, T.~Dean, J.~Kirman, and M.~Lejter.
\newblock A decision-theoretic approach to planning, perception, and control.
\newblock \emph{IEEE Expert}, 7\penalty0 (4):\penalty0 58--65, 1992.

\bibitem[Ha and Schmidhuber(2018)]{ha2018recurrent}
D.~Ha and J.~Schmidhuber.
\newblock Recurrent world models facilitate policy evolution.
\newblock In \emph{Advances in Neural Information Processing Systems}, pages
  2450--2462, 2018.

\bibitem[Kingma and Welling(2014)]{kingma2013auto}
D.~P. Kingma and M.~Welling.
\newblock Auto-encoding variational {B}ayes.
\newblock In \emph{Proceedings of the International Conference on Learning
  Representations}, 2014.

\bibitem[Igl et~al.(2018)Igl, Zintgraf, Le, Wood, and Whiteson]{igl2018deep}
M.~Igl, L.~Zintgraf, T.~A. Le, F.~Wood, and S.~Whiteson.
\newblock Deep variational reinforcement learning for {POMDP}s.
\newblock In \emph{Proceedings of the International Conference on Machine
  Learning}, pages 2117--2126, 2018.

\bibitem[Hafner et~al.(2018)Hafner, Lillicrap, Fischer, Villegas, Ha, Lee, and
  Davidson]{hafner2018planet}
D.~Hafner, T.~Lillicrap, I.~Fischer, R.~Villegas, D.~Ha, H.~Lee, and
  J.~Davidson.
\newblock Learning latent dynamics for planning from pixels.
\newblock \emph{arXiv preprint arXiv:1811.04551}, 2018.

\bibitem[Camacho and Alba(2013)]{camacho2013model}
E.~F. Camacho and C.~B. Alba.
\newblock \emph{Model predictive control}.
\newblock Springer Science \& Business Media, 2013.

\bibitem[Tassa et~al.(2018)Tassa, Doron, Muldal, Erez, Li, Casas, Budden,
  Abdolmaleki, Merel, Lefrancq, et~al.]{tassa2018deepmind}
Y.~Tassa, Y.~Doron, A.~Muldal, T.~Erez, Y.~Li, D.~d.~L. Casas, D.~Budden,
  A.~Abdolmaleki, J.~Merel, A.~Lefrancq, et~al.
\newblock Deepmind control suite.
\newblock \emph{arXiv preprint arXiv:1801.00690}, 2018.

\bibitem[Coumans and Bai(2016--2019)]{coumans2019}
E.~Coumans and Y.~Bai.
\newblock Pybullet, a python module for physics simulation for games, robotics
  and machine learning.
\newblock \url{http://pybullet.org}, 2016--2019.

\bibitem[Doya et~al.(2002)Doya, Samejima, Katagiri, and
  Kawato]{doya2002multiple}
K.~Doya, K.~Samejima, K.-i. Katagiri, and M.~Kawato.
\newblock Multiple model-based reinforcement learning.
\newblock \emph{Neural computation}, 14\penalty0 (6):\penalty0 1347--1369,
  2002.

\bibitem[Karkus et~al.(2017)Karkus, Hsu, and Lee]{karkus2017qmdp}
P.~Karkus, D.~Hsu, and W.~S. Lee.
\newblock {QMDP}-net: Deep learning for planning under partial observability.
\newblock In \emph{Advances in Neural Information Processing Systems}, pages
  4694--4704, 2017.

\bibitem[Ma et~al.()Ma, Karkus, Hsu, and Lee]{ma2019particle}
X.~Ma, P.~Karkus, D.~Hsu, and W.~S. Lee.
\newblock Particle filter recurrent neural networks.
\newblock In \emph{The Thirty-Fourth {AAAI} Conference on Artificial
  Intelligence, {AAAI}, 2020}, pages 5101--5108.

\bibitem[Hafner et~al.(2019)Hafner, Lillicrap, Ba, and
  Norouzi]{hafner2019dream}
D.~Hafner, T.~Lillicrap, J.~Ba, and M.~Norouzi.
\newblock Dream to control: Learning behaviors by latent imagination.
\newblock \emph{arXiv preprint arXiv:1912.01603}, 2019.

\bibitem[Agrawal et~al.(2016)Agrawal, Nair, Abbeel, Malik, and
  Levine]{agrawal2016learning}
P.~Agrawal, A.~V. Nair, P.~Abbeel, J.~Malik, and S.~Levine.
\newblock Learning to poke by poking: Experiential learning of intuitive
  physics.
\newblock In \emph{Advances in neural information processing systems}, pages
  5074--5082, 2016.

\bibitem[Finn and Levine(2017)]{finn2017deep}
C.~Finn and S.~Levine.
\newblock Deep visual foresight for planning robot motion.
\newblock In \emph{2017 IEEE International Conference on Robotics and
  Automation (ICRA)}, pages 2786--2793. IEEE, 2017.

\bibitem[Mnih and Teh(2012)]{mnih2012fast}
A.~Mnih and Y.~W. Teh.
\newblock A fast and simple algorithm for training neural probabilistic
  language models.
\newblock \emph{arXiv preprint arXiv:1206.6426}, 2012.

\bibitem[Pedagadi et~al.(2013)Pedagadi, Orwell, Velastin, and
  Boghossian]{pedagadi2013local}
S.~Pedagadi, J.~Orwell, S.~Velastin, and B.~Boghossian.
\newblock Local fisher discriminant analysis for pedestrian re-identification.
\newblock In \emph{Proceedings of the IEEE conference on computer vision and
  pattern recognition}, pages 3318--3325, 2013.

\bibitem[Grover and Leskovec(2016)]{grover2016node2vec}
A.~Grover and J.~Leskovec.
\newblock node2vec: Scalable feature learning for networks.
\newblock In \emph{Proceedings of the 22nd ACM SIGKDD international conference
  on Knowledge discovery and data mining}, pages 855--864, 2016.

\bibitem[Kipf et~al.(2020)Kipf, van~der Pol, and Welling]{kipf2019contrastive}
T.~Kipf, E.~van~der Pol, and M.~Welling.
\newblock Contrastive learning of structured world models.
\newblock In \emph{International Conference on Learning Representations}, 2020.

\bibitem[Karkus et~al.(2018)Karkus, Hsu, and Lee]{karkus2018particle}
P.~Karkus, D.~Hsu, and W.~S. Lee.
\newblock Particle filter networks with application to visual localization.
\newblock In \emph{Proceedings of the Conference on Robot Learning}, pages
  169--178, 2018.

\bibitem[Ma et~al.(2020)Ma, Karkus, Hsu, Lee, and Ye]{ma2020discriminative}
X.~Ma, P.~Karkus, D.~Hsu, W.~S. Lee, and N.~Ye.
\newblock Discriminative particle filter reinforcement learning for complex
  partial observations.
\newblock In \emph{International Conference on Learning Representations}, 2020.

\bibitem[Karkus et~al.(2020)Karkus, Angelova, Vanhoucke, and
  Jonschkowski]{karkus2020differentiable}
P.~Karkus, A.~Angelova, V.~Vanhoucke, and R.~Jonschkowski.
\newblock Differentiable mapping networks: Learning structured map
  representations for sparse visual localization.
\newblock \emph{arXiv preprint arXiv:2005.09530}, 2020.

\bibitem[Chung et~al.(2015)Chung, Kastner, Dinh, Goel, Courville, and
  Bengio]{chung2015recurrent}
J.~Chung, K.~Kastner, L.~Dinh, K.~Goel, A.~C. Courville, and Y.~Bengio.
\newblock A recurrent latent variable model for sequential data.
\newblock In \emph{Advances in Neural Information Processing Systems}, pages
  2980--2988, 2015.

\bibitem[LeCun et~al.(2006)LeCun, Chopra, Hadsell, Ranzato, and
  Huang]{lecun2006tutorial}
Y.~LeCun, S.~Chopra, R.~Hadsell, M.~Ranzato, and F.~Huang.
\newblock A tutorial on energy-based learning.
\newblock \emph{Predicting structured data}, 1\penalty0 (0), 2006.

\bibitem[Oord et~al.(2018)Oord, Li, and Vinyals]{oord2018representation}
A.~v.~d. Oord, Y.~Li, and O.~Vinyals.
\newblock Representation learning with contrastive predictive coding.
\newblock \emph{arXiv preprint arXiv:1807.03748}, 2018.

\bibitem[Velickovic et~al.(2019)Velickovic, Fedus, Hamilton, Li{\`o}, Bengio,
  and Hjelm]{velickovic2019deep}
P.~Velickovic, W.~Fedus, W.~L. Hamilton, P.~Li{\`o}, Y.~Bengio, and R.~D.
  Hjelm.
\newblock Deep graph infomax.
\newblock In \emph{International Conference on Learning Representations}, 2019.

\bibitem[Alemi et~al.(2016)Alemi, Fischer, Dillon, and Murphy]{alemi2016deep}
A.~A. Alemi, I.~Fischer, J.~V. Dillon, and K.~Murphy.
\newblock Deep variational information bottleneck.
\newblock \emph{arXiv preprint arXiv:1612.00410}, 2016.

\bibitem[Haarnoja et~al.(2018)Haarnoja, Zhou, Abbeel, and
  Levine]{haarnoja2018soft}
T.~Haarnoja, A.~Zhou, P.~Abbeel, and S.~Levine.
\newblock Soft actor-critic: Off-policy maximum entropy deep reinforcement
  learning with a stochastic actor.
\newblock \emph{arXiv preprint arXiv:1801.01290}, 2018.

\bibitem[Karkus et~al.(2019)Karkus, Ma, Hsu, Kaelbling, Lee, and
  Lozano-P{\'e}rez]{karkus2019dan}
P.~Karkus, X.~Ma, D.~Hsu, L.~P. Kaelbling, W.~S. Lee, and T.~Lozano-P{\'e}rez.
\newblock Differentiable algorithm networks for composable robot learning.
\newblock \emph{arXiv preprint arXiv:1905.11602}, 2019.

\bibitem[Tedrake(2009)]{tedrake2009underactuated}
R.~Tedrake.
\newblock Underactuated robotics: Learning, planning, and control for efficient
  and agile machines course notes for mit 6.832.
\newblock \emph{Working draft edition}, 3, 2009.

\bibitem[Barth-Maron et~al.(2018)Barth-Maron, Hoffman, Budden, Dabney, Horgan,
  Tb, Muldal, Heess, and Lillicrap]{barth2018distributed}
G.~Barth-Maron, M.~W. Hoffman, D.~Budden, W.~Dabney, D.~Horgan, D.~Tb,
  A.~Muldal, N.~Heess, and T.~Lillicrap.
\newblock Distributed distributional deterministic policy gradients.
\newblock \emph{arXiv preprint arXiv:1804.08617}, 2018.

\bibitem[Russakovsky et~al.(2015)Russakovsky, Deng, Su, Krause, Satheesh, Ma,
  Huang, Karpathy, Khosla, Bernstein, et~al.]{russakovsky2015imagenet}
O.~Russakovsky, J.~Deng, H.~Su, J.~Krause, S.~Satheesh, S.~Ma, Z.~Huang,
  A.~Karpathy, A.~Khosla, M.~Bernstein, et~al.
\newblock Imagenet large scale visual recognition challenge.
\newblock \emph{International journal of computer vision}, 115\penalty0
  (3):\penalty0 211--252, 2015.

\bibitem[Li et~al.(2018{\natexlab{a}})Li, Lee, and Hsu]{li2018push}
J.~K. Li, W.~S. Lee, and D.~Hsu.
\newblock Push-net: Deep planar pushing for objects with unknown physical
  properties.
\newblock In \emph{Robotics: Science and Systems}, volume~14, pages 1--9,
  2018{\natexlab{a}}.

\bibitem[Li et~al.(2018{\natexlab{b}})Li, Wu, Tedrake, Tenenbaum, and
  Torralba]{li2018learning}
Y.~Li, J.~Wu, R.~Tedrake, J.~B. Tenenbaum, and A.~Torralba.
\newblock Learning particle dynamics for manipulating rigid bodies, deformable
  objects, and fluids.
\newblock \emph{arXiv preprint arXiv:1810.01566}, 2018{\natexlab{b}}.

\end{thebibliography}
\end{document}